%% file: iclr2027_conference.tex
\documentclass{article} 
\usepackage{iclr2027_conference,times}

\input{math_commands.tex}

\usepackage{hyperref}
\usepackage{graphicx}
\usepackage{booktabs}
\usepackage{multirow}
\usepackage[table]{xcolor}
\usepackage{amsmath, amsfonts}
\usepackage{subcaption}
\usepackage{makecell}
\usepackage{algorithm}
\usepackage{algorithmic}
\usepackage{url}
\usepackage{wrapfig}

\title{CoDrive: Cross-Vehicle World-Consistent Video Generation with Precise Trajectory Control for Cooperative Driving}

\author{
Yu Meng*\dag \textsuperscript{\rm 2},
Baining Zhao*\dag \textsuperscript{\rm 2},
Junta Wu\textsuperscript{\rm 1},
Tengfei Wang\ddag \textsuperscript{\rm 1},
Rongze Tang\textsuperscript{\rm 3},
Haiyu Zhang\textsuperscript{\rm 1}, \\
\textbf{Wenqiang Sun}\textsuperscript{\rm 1}, 
\textbf{Chen Gao\ddag} \textsuperscript{\rm 2 3},
\textbf{Zhibo Chen}\textsuperscript{\rm 3},
\textbf{Xinlei Chen}\textsuperscript{\rm 2}, 
\textbf{Yong Li}\textsuperscript{\rm 2},
\textbf{Xiao-Ping Zhang}\textsuperscript{\rm 2}, \\
\textbf{Chunchao Guo\ddag} \textsuperscript{\rm 1} \\
\textsuperscript{\rm 1}Tencent Hunyuan,
\textsuperscript{\rm 2}Tsinghua University,
\textsuperscript{\rm 3}Zhongguancun Academy
}

\iclrfinalcopy 
\begin{document}

\maketitle

{
  \let\thefootnote\relax
  \footnotetext{* Equal Contribution. \dag\ Work done during internship at Tencent Hunyuan.}
  \footnotetext{\ddag\ Corresponding Authors.}
}

\begin{figure*}[h]
    \centering
        \includegraphics[width=\linewidth]{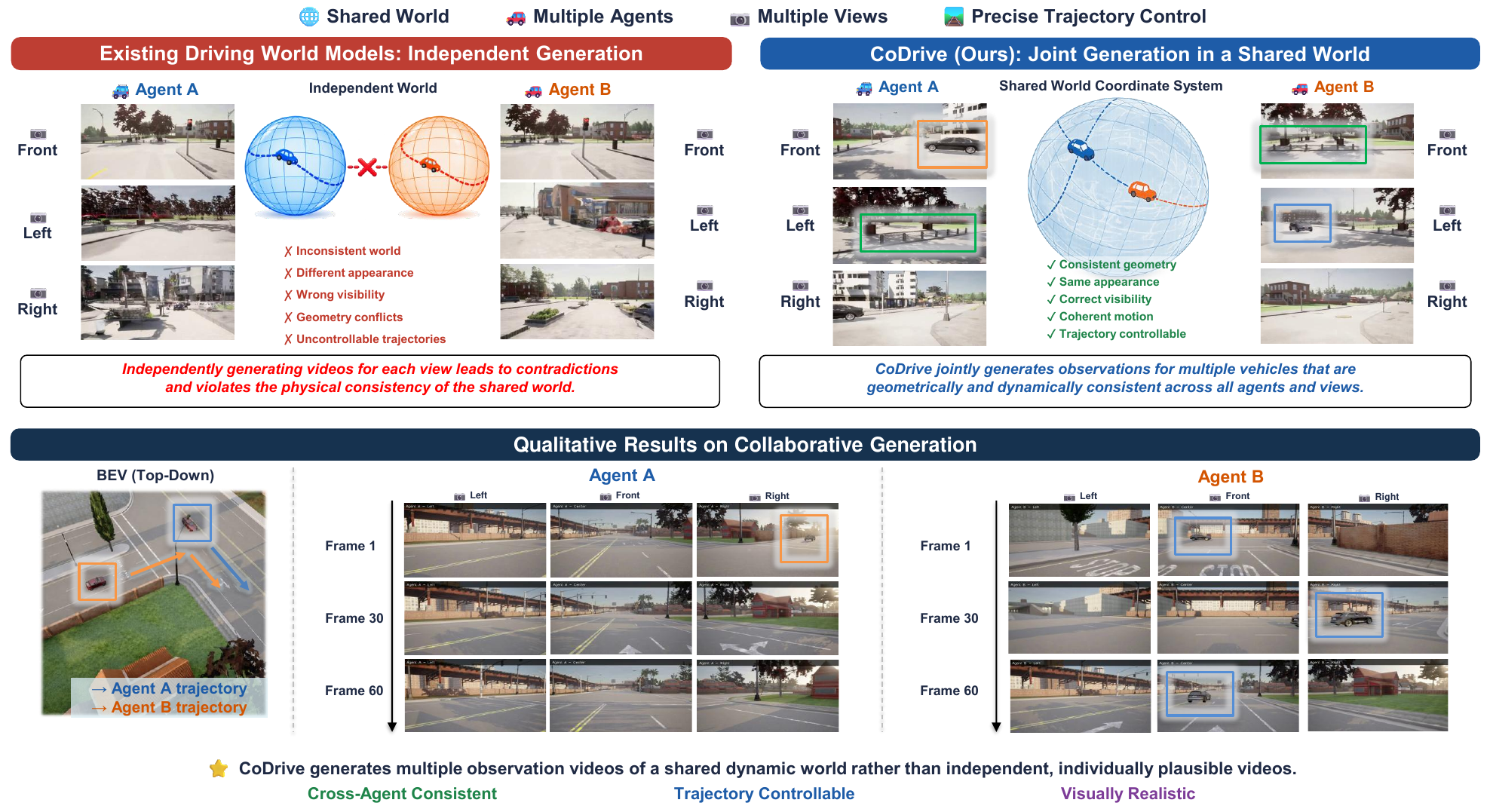}
    \caption{\textbf{CoDrive enables controllable, cross-vehicle consistent driving video generation}.
    }
\label{fig:teaser}
\end{figure*}

\input{sections/0-abstract}
\input{sections/1-introduction}
\input{sections/2-related-work}
\input{sections/3-method}
\input{sections/4-benchmark}
\input{sections/5-experiment}
\input{sections/6-conclusion}

\bibliography{iclr2027_conference}
\bibliographystyle{iclr2027_conference}
\clearpage
\appendix
\input{sections/7-appendix}

\end{document}

%% file: math_commands.tex
\usepackage{amsmath,amsfonts,bm}

\def\eqref#1{equation~\ref{#1}}

\def\1{\bm{1}}

\DeclareMathAlphabet{\mathsfit}{\encodingdefault}{\sfdefault}{m}{sl}
\SetMathAlphabet{\mathsfit}{bold}{\encodingdefault}{\sfdefault}{bx}{n}



%% file: sections/0-abstract.tex
\begin{abstract}

Real-world driving is inherently multi-agent, yet most existing driving world models generate observations from a single ego vehicle.
Independently extending them to multiple vehicles does not ensure that different agents observe a consistent shared world.
We present CoDrive, a cross-vehicle, multi-view driving video generation framework that jointly generates observations of vehicles sharing the same dynamic scene with precise camera-trajectory control.
CoDrive interleaves local self-attention, which models spatiotemporal dependencies among the views of each vehicle, with global self-attention, which enables information exchange and consistency modeling across vehicles.
To explicitly encode their spatial relationships, all camera trajectories are represented in a shared world coordinate system and injected into the attention layers through projective relative positional encoding.
We further adopt a progressive mixed-task training strategy that combines large-scale real-world single-agent data with synthetic cross-agent interaction data, allowing the model to benefit from real-world appearance distributions while learning cross-agent consistency from simulation.
For systematic evaluation, we introduce CoDrive-Bench, a benchmark covering real and synthetic multi-vehicle scenarios and evaluating trajectory controllability, scene geometry consistency, and instance-level consistency.
Experiments show that CoDrive improves trajectory controllability and cross-agent geometric and instance consistency while maintaining competitive visual quality.
Project page: \url{https://codrive-project-page.github.io/}.

\end{abstract}

%% file: sections/1-introduction.tex
\section{Introduction}

Recent advances in video generation have enabled generative models to produce visually realistic and temporally coherent videos, opening up the possibility of learning world models directly from large-scale visual data.
By modeling how a scene evolves over time under different conditions, video world models provide a scalable approach to simulating complex dynamic environments.
Autonomous driving is a particularly important application of this paradigm, as a capable driving world model could support data generation, closed-loop evaluation, rare-event simulation, and the development of perception and planning systems.
Existing driving world models have made substantial progress in visual fidelity, controllability, long-horizon generation, multimodal prediction, and physical plausibility.
However, most existing methods model the driving world from the perspective of a \emph{single} ego vehicle.

Real-world traffic environments are inherently multi-agent.
Multiple vehicles simultaneously observe and interact with the same dynamic scene from different locations, orientations, and fields of view.
A world model for such environments should therefore generate not only individually plausible observations, but observations that are mutually consistent with a single shared world.
For example, when two vehicles observe the same road user, intersection, or traffic event, the generated observations should agree on the identity, location, appearance, motion, and visibility of the shared content.
Independently generating the observation of each vehicle can easily lead to contradictions: an object may appear in only one view, follow incompatible trajectories across agents, or occupy geometrically inconsistent positions.
Such inconsistencies limit the use of conventional single-agent generators for cooperative perception, multi-vehicle planning, and interactive traffic simulation.

In this work, we introduce CoDrive, a generative framework that jointly models multi-view observations of vehicles interacting in a shared driving scene.
The task naturally involves two levels of consistency: coherence among the views of each vehicle and agreement across vehicles.
Accordingly, we interleave local self-attention for intra-agent multi-view modeling with global self-attention for cross-agent information exchange.
To provide explicit geometric guidance, we express all camera trajectories in a common world coordinate system and inject their relative geometry into the attention layers.
This enables cross-vehicle interaction to depend on the physical relationship between cameras rather than appearance similarity alone.
Through multi-stage progressive mixed training on diverse data, CoDrive demonstrates consistent generation capabilities in cross-agent scenarios.

To enable systematic evaluation, we introduce CoDrive-Bench, a dedicated benchmark for cross-agent, multi-view driving world models.
Beyond conventional video quality, CoDrive-Bench evaluates whether generated observations follow the commanded trajectories, reconstruct consistent scene geometry, and place shared vehicle instances consistently across agents.

Our contributions are summarized as follows:
\begin{itemize}
    \item We introduce CoDrive, a framework for jointly generating multi-view observations of two vehicles in a shared dynamic scene with explicit camera-trajectory control.
    \item We construct CoDrive-Bench, a comprehensive benchmark that quantitatively evaluates controllability and cross-agent consistency across vehicles.
    \item We conduct experiments on CoDrive-Bench and demonstrate that CoDrive improves trajectory controllability and cross-agent consistency over the evaluated baselines while retaining competitive visual quality.
\end{itemize}

%% file: sections/2-related-work.tex
\section{Related Work}

\paragraph{Video World Models.}
In recent years, large-scale latent-space video diffusion pretraining~\citep{zheng2024opensorademocratizingefficientvideo,rombach2022high,kong2025hunyuanvideosystematicframeworklarge,wan2025wanopenadvancedlargescale,hong2022cogvideolargescalepretrainingtexttovideo} has endowed video diffusion models with rich world knowledge, giving rise to emergent capabilities for understanding and predicting the visual world.
As a result, they are able to generate video sequences with high visual fidelity, temporal coherence, and physical plausibility.
Moreover, a growing body of work has introduced various conditioning signals into the world generation process, including text~\citep{mao2025yume15textcontrolledinteractiveworld}, camera poses~\citep{he2025cameractrl,ren2025gen3c,zhang2026worldstereo,sun2025worldplay,zhu2026sanawmefficientminutescaleworld,wang2026worldcompass}, and actions~\citep{he2026matrixgame20opensourcerealtime,wang2026matrixgame30realtimestreaming,tang2026hunyuangamecraft2instructionfollowinginteractivegame,nvidia2026cosmos3omnimodalworld}, thereby enabling video generation to be controlled through multiple modalities and allowing more precise manipulation of the visual worlds predicted by the models.

\paragraph{Multi-Agent World Models.}
With the rapid advancement of video world models, recent studies have begun to explore joint world modeling shared across multiple agents.
Existing approaches either control multiple agents within a single viewpoint~\citep{pondaven2026actionpartymultisubjectactionbinding,zhu2026incantationnaturallanguageaction} or jointly generate the egocentric views of multiple agents~\citep{savva2026solarisbuildingmultiplayervideo,hu2026metaworldscalingmultiagentvideo,wu2026multiworldscalablemultiagentmultiview,sun2026prismaworldcameracontrollablemultiagentvideo,liu2026gammaworldgenerativemultiagentworld,hu2026miramultiplayerinteractiveworldmodels}, thereby achieving spatiotemporal consistency across agents.
However, most of these methods are confined to game environments such as Minecraft and It Takes Two.
Multi-agent world models for real-world applications such as autonomous driving remain largely underexplored.

\paragraph{Driving World Models.}
Autonomous driving represents an important application domain for video world models with powerful world-simulation capabilities.
Early studies~\citep{hu2023gaia1generativeworldmodel,yang2024genad} trained video generation models on driving videos to simulate driving scenarios.
Subsequent works further advanced driving world models in terms of controllability~\citep{li2023drivingdiffusion,lu2023wovogen,wang2024drivedreamer,hassan2025gem,jiang2024diveditbasedvideogeneration,gao2024vista}, resolution~\citep{wu2024drivescapehighresolutioncontrollablemultiview,gao2025magicdrive-v2}, output modalities~\citep{li2025uniscene,guo2025genesis,li2025omninwm}, generation horizon~\citep{chen2025unimlvgunifiedframeworkmultiview,zhang2025epona}, physical consistency~\citep{yang2026geniedrive,zhou2026physicallyconsistentdrivingvideo}, and scene plausibility~\citep{wang2024drivingdojo,yan2026causaldriverealtimecausalworld,zhou2025safemvdrive}.
However, most of these approaches generate the driving world from a single ego-vehicle perspective.
\citet{tao2026v2vcrafterconsistentstreetviewimage} explores multi-agent street view generation, but focuses primarily on image generation task rather than video world modeling.
\citet{zhu2026shareversemultiagentconsistentvideo} explores dual-vehicle world video generation, but focuses only on simulated environments and does not provide a systematic evaluation of cross-vehicle controllability and consistency.

%% file: sections/3-method.tex
\section{CoDrive}~\label{sec:method}\vspace{-2.5em}

\subsection{Model Architecture}\label{sec:model_arch}

\begin{figure*}[t!]
    \centering
        \includegraphics[width=\linewidth]{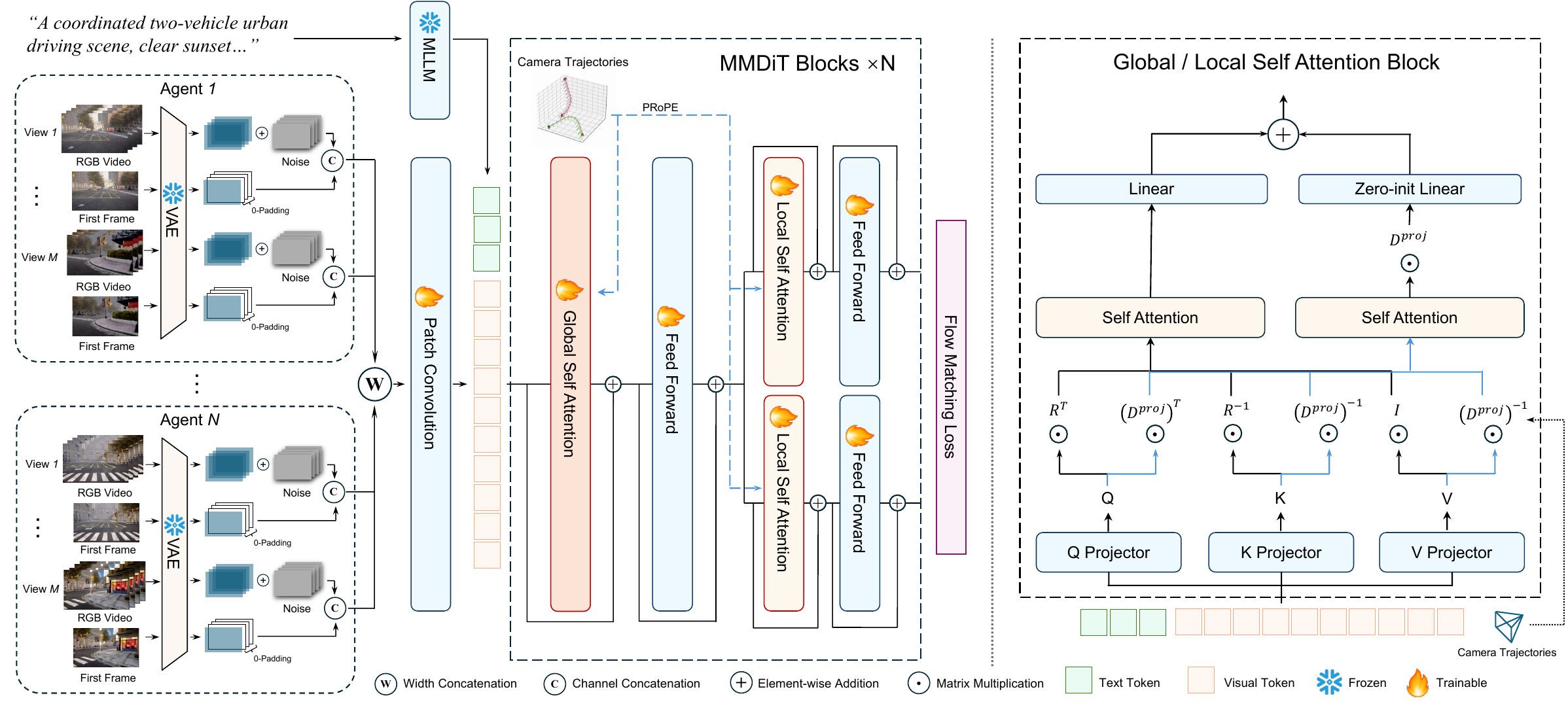}
    \caption{\textbf{Model architecture and training framework overview.} Given text prompt, reference frames, and camera poses from multiple agents and views, CoDrive encodes each video and jointly denoises them with a pretrained video DiT. Interleaved local self-attention models intra-agent spatial-temporal dependencies across multiple views, while global self-attention enables information exchange and consistency modeling across vehicles. Camera poses are expressed in a shared world coordinate system and injected into both attention modules through projective relative positional encoding (PRoPE), providing explicit geometric guidance for cross-view and cross-agent alignment.}
\label{fig:framework}
\end{figure*}

As shown in Figure~\ref{fig:framework}, our generator is built on a pretrained latent video diffusion transformer. A $3$D VAE encodes video into a compact latent and a diffusion transformer denoises it under a text prompt and a
reference frame.
We retain the pretrained backbone and extend it to jointly generate observations from multiple vehicles through two complementary mechanisms: cross-agent information exchange and explicit shared-camera geometry.

\paragraph{Cross-agent multi-view formulation.}
We model the shared scene as the set of views captured by $N$ vehicles, each equipped with $M$ forward-facing cameras, giving $V=NM$ views in total.
Let $\mathbf{x}_v$ denote the video of view $v$. We encode each view independently with the frozen VAE encoder $\mathcal{E}$ and concatenate the per-view latents along the width axis:
\begin{equation}
  \mathbf{z}_v = \mathcal{E}(\mathbf{x}_v)\in\mathbb{R}^{T\times H\times W\times C},
  \qquad
  \mathbf{z} = \big[\,\mathbf{z}_1 \,\Vert\, \mathbf{z}_2 \,\Vert\, \cdots \,\Vert\, \mathbf{z}_V\,\big]_{\text{w}}
  \in\mathbb{R}^{\,T\times H\times (V\!\cdot\!W)\times C},
  \label{eq:wide_latent}
\end{equation}
where $[\cdot\Vert\cdot]_{\text{w}}$ is concatenation along width and $(T,H,W,C)$
are the per-view latent dimensions. The transformer operates on the token
sequence of this wide latent $\mathbf{z}$.

\paragraph{Interleaved Global-Local Self-Attention.}
Cross-agent consistency first requires each vehicle’s own multi-view observations to remain spatiotemporally coherent.
We therefore use Local Self-Attention to model dependencies among the views belonging to the same vehicle.
Given the video tokens associated with one agent, this module captures their spatiotemporal dependencies across different views.
Specifically, for each agent, the tokens from all views are sequentially fed into the Local Self-Attention module, and the corresponding outputs are concatenated to form a unified representation:
\begin{equation}
  \mathrm{Attn}_{\text{local}}(\mathbf{z}) =
  \big\Vert_{\,n=1}^{\,N}\ \mathrm{Attn}\!\big(\mathbf{z}^{(n)}\big),
  \qquad
  \mathbf{z}^{(n)}=\big[\mathbf{z}_{(n-1)M+1}\Vert\cdots\Vert\mathbf{z}_{nM}\big]_{\text{w}},
\end{equation}
where $\mathbf{z}^{(n)}$ gathers the $M$ views of agent $n$.

However, local attention alone cannot make different agents agree on the world they share.
We therefore apply Global Self-Attention to the wide latent spanning all agents, allowing tokens from one vehicle to attend to observations from other vehicles and reconcile shared scene content:
\begin{equation}
  \mathrm{Attn}_{\text{global}}(\mathbf{z}) =\mathrm{Attn}\!\big(\big[\,\mathbf{z}_1 \,\Vert\, \mathbf{z}_2 \,\Vert\, \cdots \,\Vert\, \mathbf{z}_V\,\big]_{\text{w}}\big).
\end{equation}

We interleave local and global attention across transformer blocks. Local blocks preserve intra-agent multi-view structure, while global blocks progressively propagate information across vehicles, allowing cross-agent consistency to emerge without replacing per-agent representations with fully global attention at every layer.

\paragraph{Global Camera Geometry Injection.}
Global Self-Attention allows tokens from different agents to interact, but does not by itself specify how their cameras are positioned in the shared scene.
Therefore, we express all cameras relative to one common reference camera, rather than normalizing each agent independently.
This places all views in a shared geometric gauge, so that cross-agent attention can depend on their physical camera relationships rather than on appearance correspondence alone.
Let $a=(t,v)$ index view $v$ at latent time $t$, and let $\mathbf E_a={}^{C_a}\mathbf T_W\in\mathrm{SE}(3)$ be its world-to-camera transform.
Using $a_0=(0,0)$ as the shared reference, we define
\begin{equation}
  \widetilde{\mathbf E}_a
  =\mathbf E_a\mathbf E_{a_0}^{-1}
  ={}^{C_a}\mathbf T_{C_{a_0}},
  \label{eq:global_gauge}
\end{equation}
which maps coordinates from the reference-camera frame to camera $a$.
Thus, $\widetilde{\mathbf E}_a$ is a relative camera transform in one shared canonical gauge, rather than an absolute world-to-camera pose.

We inject these poses using an implementation of projective relative positional encoding (PRoPE) adapted from~\citep{li2025prope,sun2025worldplay}.
After normalizing the focal lengths and removing the principal-point entries, we construct an invertible homogeneous intrinsic transform
\begin{equation}
  \widehat{\mathbf K}_a
  =
  \begin{bmatrix}
    \operatorname{diag}(\kappa_a^x,\kappa_a^y,1) & \mathbf 0\\
    \mathbf 0^\top & 1
  \end{bmatrix}
  \in\mathrm{GL}(4),
  \qquad
  \mathbf P_a=\widehat{\mathbf K}_a\widetilde{\mathbf E}_a
  \in\mathrm{GL}(4).
  \label{eq:prope_matrix}
\end{equation}
Here $\mathbf P_a$ is an invertible homogeneous feature transform, not a physical perspective projection involving depth division.

The camera tensors have shapes $\widetilde{\mathbf E}\in\mathbb R^{B\times T\times V\times4\times4}$ and $\mathbf K\in\mathbb R^{B\times T\times V\times3\times3}$, and are broadcast to the image tokens belonging to the corresponding $(t,v)$.
Specifically, for an attention head of dimension $d_h=4R$, PRoPE divides each learned query, key, and value into $R$ four-dimensional channel groups and applies the block-diagonal transform
$\mathcal P_a=\mathbf I_R\otimes\mathbf P_a\in\mathbb R^{d_h\times d_h}$:
\begin{equation}
  \mathbf q'_a=\mathcal P_a^\top\mathbf q_a,\qquad
  \mathbf k'_a=\mathcal P_a^{-1}\mathbf k_a,\qquad
  \mathbf v'_a=\mathcal P_a^{-1}\mathbf v_a .
  \label{eq:prope_qkv}
\end{equation}
$\mathbf q_a$, $\mathbf k_a$ and $\mathbf v_a$ are the learned image Q/K/V features after their linear projections and Q/K normalization.
As shown in the right part of Figure~\ref{fig:framework}, the transformed features are processed by an additional attention branch, whose image output is mapped back by $\mathcal P_a$ and added to the pretrained attention output through a zero-initialized projection. For image tokens $i$ and $j$, the resulting attention logit is
\begin{equation}
  \left\langle\mathbf q'_i,\mathbf k'_j\right\rangle
  =
  \mathbf q_i^\top
  \mathcal P_i\mathcal P_j^{-1}
  \mathbf k_j,
  \qquad
  \mathbf P_i\mathbf P_j^{-1}
  =
  \widehat{\mathbf K}_i
  \mathbf E_i\mathbf E_j^{-1}
  \widehat{\mathbf K}_j^{-1}.
  \label{eq:prope_relative}
\end{equation}
The shared reference $\mathbf E_{a_0}$ therefore cancels, while $\mathbf E_i\mathbf E_j^{-1}={}^{C_i}\mathbf T_{C_j}$ is the relative extrinsic transform from camera $j$ to camera $i$.
The full expression additionally contains the two camera intrinsics and is therefore an intrinsic-calibrated relative projective operator, rather than a pure relative rigid transform.
It supplies intra-agent camera geometry in local-attention blocks and cross-agent camera geometry in global-attention blocks.
The invariance to the original world-coordinate gauge and the complete derivation are provided in Appendix~\ref{sec:appendix-prope}.
Intuitively, the resulting attention interaction depends on the calibrated relative geometry between the source and target cameras, while remaining invariant to the arbitrary choice of the shared world-coordinate gauge.

\paragraph{Training Objective.}
Following the base model, we train with a flow-matching objective. Given the clean wide latent $\mathbf{z}$ and Gaussian noise $\boldsymbol{\epsilon}\!\sim\!\mathcal{N}(\mathbf{0},\mathbf{I})$, we draw a flow time $\tau\!\in\![0,1]$ and form the linear interpolant
\begin{equation}
\mathbf{z}_\tau = (1-\tau)\,\mathbf{z} + \tau\,\boldsymbol{\epsilon},
\end{equation}
whose constant velocity along the path is $\boldsymbol{\epsilon}-\mathbf{z}$.
The transformer $\mathbf{v}_\theta$ regresses this velocity from the noised latent, conditioned on the text and the reference frame conditions $c$, and the per-view cameras $\{\mathbf{P}_v\}$:
\begin{equation}
\mathcal{L} = \mathbb{E}_{\tau,\,\mathbf{z},\,\boldsymbol{\epsilon}}
\Big[\big\lVert\, \mathbf{v}_\theta(\mathbf{z}_\tau,\tau,c,\{\mathbf{P}_v\})
 - (\boldsymbol{\epsilon}-\mathbf{z})\big\rVert_2^2 \Big].
\label{eq:flow_matching}
\end{equation}

\subsection{Training Recipe}
We carefully design a multi-stage progressive training pipeline.
First, we train the base model on large-scale single-view driving videos using a text-and-image-to-video (TI2V) objective to facilitate domain adaptation.
Next, we train the model on driving videos annotated with camera poses to enable controllable camera motion during generation.
In the third stage, we further introduce single-vehicle, multi-view data with camera-pose annotations to increase the number of views supported by the model.
In the fourth stage, the model is trained on multi-vehicle, multi-view data to establish spatial consistency across vehicles.
Across these stages, we progressively increase the training resolution and combine real-world and simulated data according to the supervision available at each stage.

Moreover, since real-world multi-vehicle data are scarce, we introduce an additional mixed-task fine-tuning stage.
During this stage, the model is jointly trained on dual-agent video generation using synthetic data and single-agent video generation using real-world data.
This mixed-task formulation is designed to retain cross-agent supervision from simulation while exposing the model to the appearance and motion distributions of real-world driving scenes.

\subsection{Dataset Construction}\label{sec:method-data}

Our training data include multiple open-source autonomous driving datasets~\citep{caesar2020nuscenes,sun2020waymoscalability,yang2024genad,xiao2021pandaset,covla_wacv2025} collected in real-world environments.
We use a large vision-language model that has been trained with reinforcement learning on captioning tasks~\citep{wu2025hunyuanvideo15technicalreport} to generate textual annotations for these video data.
However, real-world multi-agent data are relatively scarce and costly to collect.
To alleviate this limitation, we develop a synthetic video data collection pipeline in the CARLA~\citep{dosovitskiy2017carla} simulator to capture interactions among dual vehicles.

The synthetic data are collected under diverse scene configurations and weather conditions, covering a wide range of multi-vehicle interaction patterns, including same-direction following, oncoming encounters, merging, cut-ins, and side-by-side driving.
Specifically, the dataset encompasses eight map scenarios, eleven weather conditions, and six vehicle interaction modes.
To further enhance the realism and diversity of the simulated data, we introduced random perturbations into vehicle motions and added task-irrelevant background traffic as distractors.
Finally, we curate 15K high-quality video clips of two-vehicle interactions from simulated environments, along with their corresponding camera pose data, for training.
Further details of data collection are in Appendix~\ref{sec:appendix-data}.

%% file: sections/4-benchmark.tex
\section{CoDrive-Bench}\label{sec:bench}

For cross-agent, multi-view driving world models, the generated content must maintain spatial consistency across agents.
However, quantitatively evaluating such consistency remains challenging.
Most existing video generation benchmarks and metrics are designed for single-agent ego videos, making them unsuitable for effectively assessing cross-agent video generation.

To address this gap, we introduce CoDrive-Bench, a benchmark suite for the quantitative evaluation of cross-agent driving world models.
CoDrive-Bench includes cross-agent data collected from both real-world and simulated environments, covering diverse weather conditions, driving scenarios, and vehicle interaction patterns.
It evaluates three complementary dimensions: camera-trajectory controllability, cross-agent scene consistency, and cross-agent instance consistency.

\subsection{Data Collection}

CoDrive-Bench contains 200 distinct two-vehicle interaction scenarios together with their corresponding textual descriptions. For each vehicle, the benchmark provides RGB videos captured from three camera views, along with a camera pose trajectory describing the vehicle’s motion. In total, the benchmark includes 1,200 RGB videos and 400 trajectories. Among the 200 scenarios, 100 are real-world scenarios and 100 are synthetic scenarios.

\begin{figure}[t]
    \centering

    \begin{subfigure}[t]{0.32\linewidth}
        \centering
        \includegraphics[width=\linewidth]{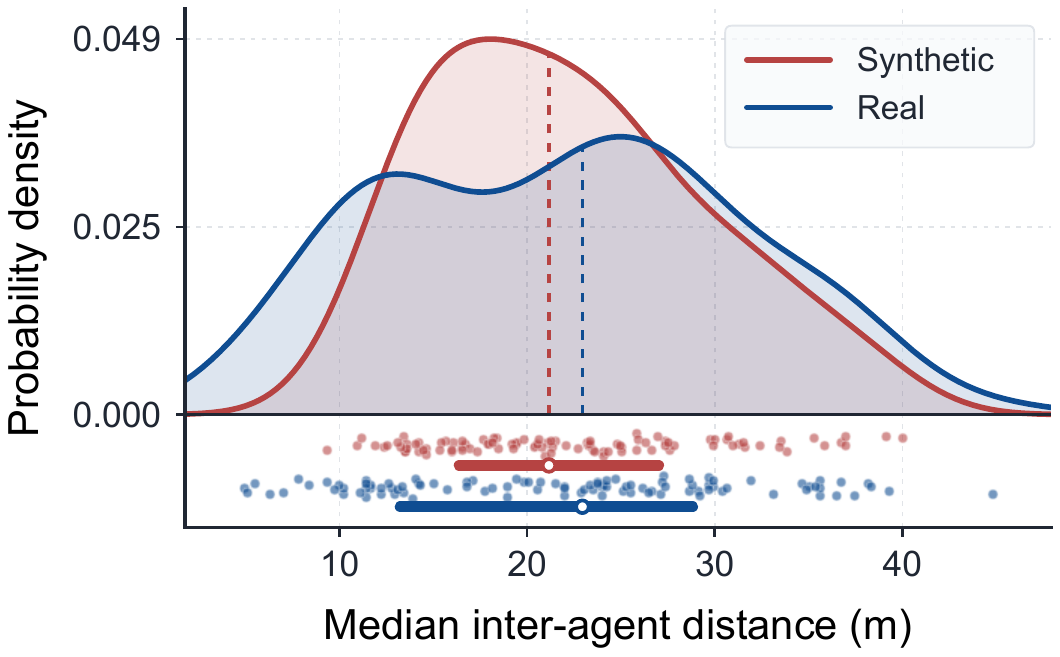}
        \caption{Agent distance distribution.}
        \label{fig:sub_a}
    \end{subfigure}
    \hfill
    \begin{subfigure}[t]{0.32\linewidth}
        \centering
        \includegraphics[width=\linewidth]{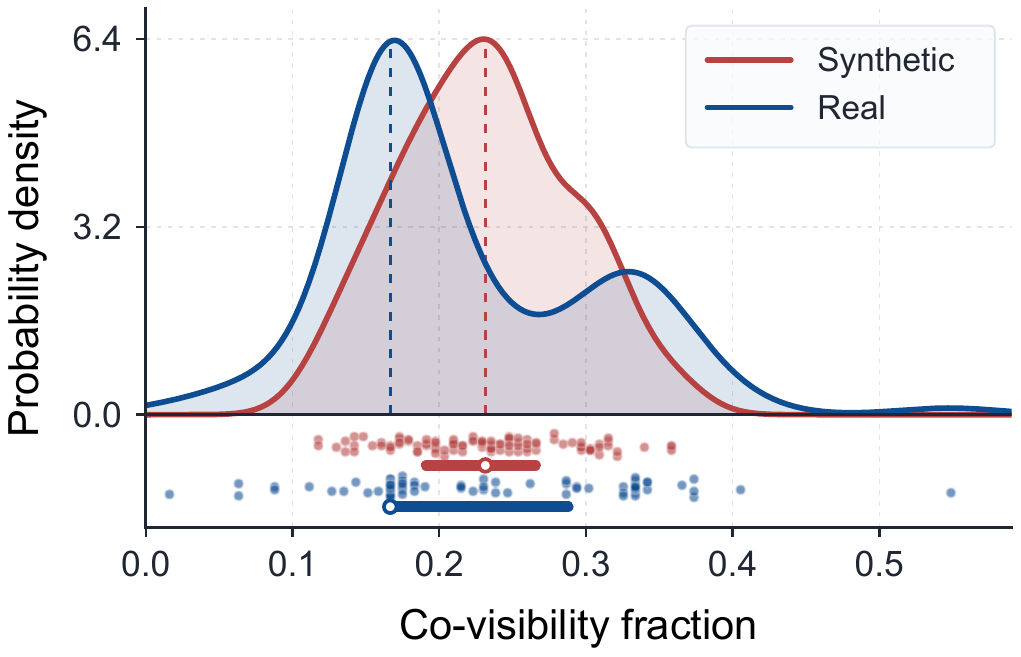}
        \caption{Co-visibility distribution.}
        \label{fig:sub_b}
    \end{subfigure}
    \hfill
    \begin{subfigure}[t]{0.32\linewidth}
        \centering
        \includegraphics[width=\linewidth]{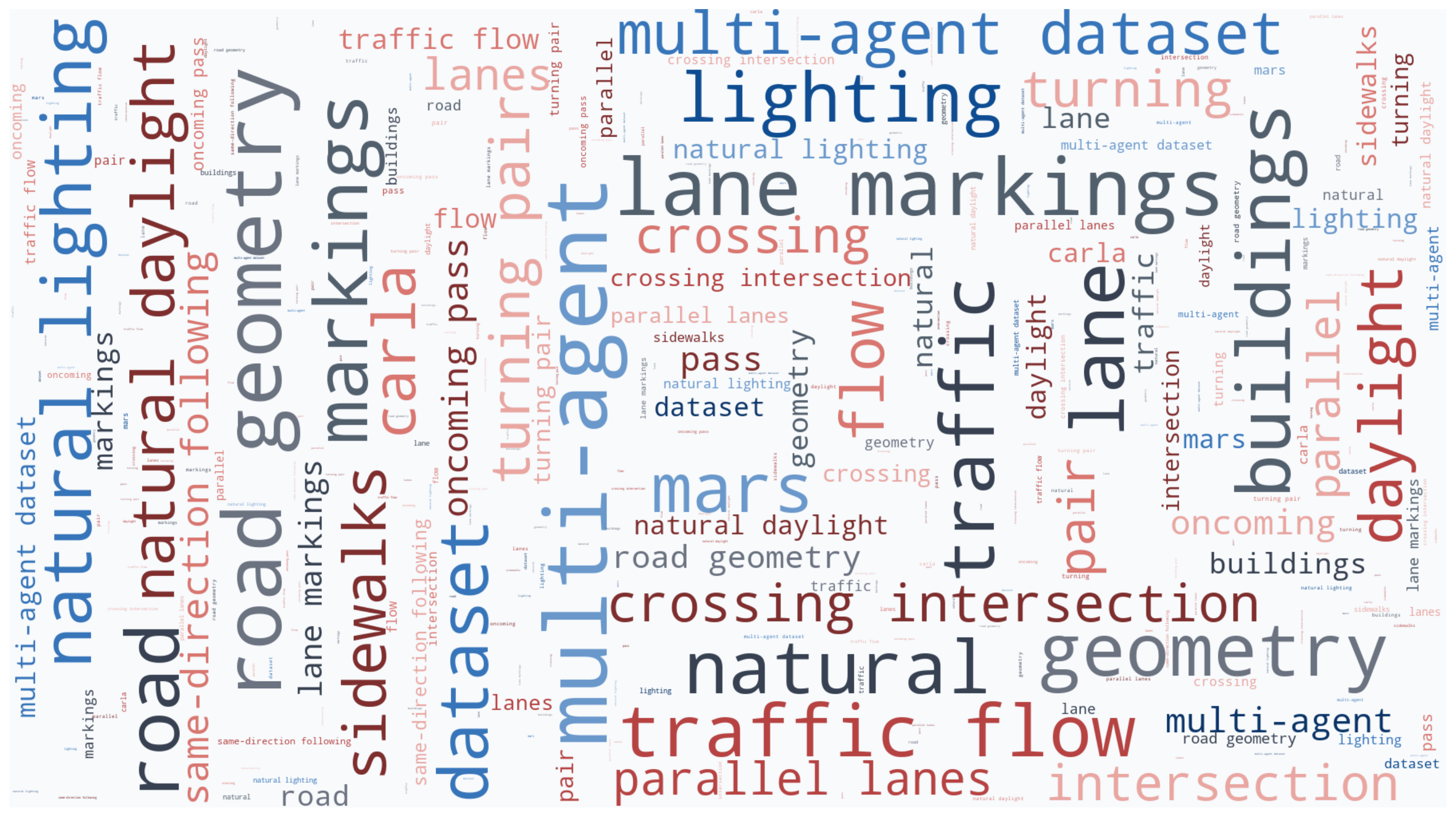}
        \caption{Prompt word cloud.}
        \label{fig:sub_c}
    \end{subfigure}

    \caption{\textbf{CoDrive-Bench data statistics.} (a) Distribution of inter-agent distances in the dataset. (b) Distribution of co-visibility between the two agents.
 (c) Word cloud of text prompts.}
    \label{fig:bench-statistics}
\end{figure}

The real-world data are drawn from OpenMars~\citep{li2024multiagent}, an open-source multi-agent, multi-view interactive video dataset, while the synthetic scenarios are collected using the pipeline described in Section~\ref{sec:method-data}. As illustrated in Fig.~\ref{fig:bench-statistics}, during data filtering, we remove video clips in which the two vehicles are too far apart or share too little overlapping field of view. We also ensure that the selected data are as diverse as possible in terms of weather conditions, scene types, and vehicle interaction patterns.

\subsection{Evaluation Metrics}

Our metrics address three complementary questions: (1) whether each generated video follows its commanded camera trajectory, (2) whether different agents reconstruct the same static scene, and (3) whether participating vehicles appear consistently at their geometrically expected locations.

\paragraph{Controllability.}
A driving world model should accurately follow the user-specified camera trajectory while preserving realistic scene dynamics.
Following DrivingGen~\citep{zhou2026drivinggen}, we estimate the camera trajectory from each generated video using monocular depth estimation together with visual SLAM.
We then evaluate trajectory accuracy using Average Displacement Error (ADE) and Dynamic Time Warping (DTW).
ADE measures the average spatial deviation between the generated and target trajectories, whereas DTW evaluates trajectory similarity under temporal misalignment, making it more robust to variations in motion speed.

\paragraph{Scene Consistency.}
Trajectory control alone does not ensure that different vehicles produce geometrically compatible observations under the supplied camera conditions.
We therefore evaluate geometric agreement between the reconstructed scenes of different agents from both three-dimensional and image-space perspectives.
Specifically, we compute Chamfer Distance (CD) between cross-agent point clouds reconstructed in the commanded world frame, measuring their geometric agreement under the supplied camera trajectories.
We further introduce Relative Reprojection Depth Error (RE), which measures cross-view depth consistency by reprojecting static scene points between different agents.
Both metrics are computed after removing dynamic foreground objects, ensuring that the evaluation focuses on the shared static environment rather than independently moving vehicles.

\paragraph{Instance Consistency.}
Scene-level geometry does not by itself guarantee that interacting vehicles are rendered consistently across agents.
We therefore evaluate whether the other participating vehicle appears along the image-space trajectory implied by the known cross-agent geometry.
Since both agents are defined in a shared world coordinate frame, the target vehicle's location is directly determined by the observing agent's camera extrinsics, eliminating the need for object annotations.
We therefore place a vehicle-sized bounding box at the corresponding location and evaluate whether the observer's video contains a vehicle at its projected image position.
Soft Hit Rate (Hit) measures whether a persistent vehicle track aligns with the geometrically predicted image trajectory by jointly considering its temporal coverage, spatial deviation, and depth consistency.
Because the prediction inherently captures the target's motion, vehicles rendered with incorrect velocity or heading deviate from the expected trajectory and are penalized, whereas transient false positives are naturally suppressed.
3D Localization Error (LE) measures the Euclidean distance, in meters, between the back-projected position of the detected vehicle and its ground-truth position, conditioned on successful rendering and detection.

Detailed metric definitions, computation procedures, and analysis are provided in Appendix~\ref{sec:appendix-bench-metrics}.

%% file: sections/5-experiment.tex
\section{Experiments}\label{sec:exp}

\paragraph{Implementation Details.} CoDrive is built upon HunyuanVideo-1.5, a pretrained video generation backbone.
Although the formulation in Section~\ref{sec:method} supports $N$ vehicles, our current implementation and evaluation use $N=2$ vehicles, each equipped with three views, following the format of the available real-world multi-vehicle data.
The training process consists of five stages and uses approximately 900K video clips in total.
More detailed information on the training procedure and dataset is provided in Appendices~\ref{sec:appendix-data} and \ref{sec:appendix-training}.

\paragraph{Baselines.}
We compare CoDrive with:
(1) general-purpose video generation models, including Wan2.2-I2V-A14B and HunyuanVideo-1.5, (2) driving-specific world models, represented by MagicDrive-V2 and Cosmos3-Nano, (3) multi-vehicle driving scene generation model for synthetic driving scenarios, represented by ShareVerse.
For each baseline, we provide the available first-frame and camera-control conditions in the format supported by that method.

\paragraph{Evaluation Protocol.}
We conduct experimental evaluations on CoDrive-Bench and report the Controllability, Scene Consistency, and Instance Consistency metrics. In addition, we report FVD to measure the visual quality. 
None of the benchmark data is included in our training.

\subsection{Main Results}

\begin{table*}[t]
    \centering
    \caption{\textbf{Quantitative comparison of video generation methods.}
    $\uparrow$ indicates that higher values are better, while
    $\downarrow$ indicates that lower values are better.}
    \label{tab:quantitative-comparison}
    \resizebox{0.95\textwidth}{!}{%
    \begin{tabular}{lcccccccc}
        \toprule
        &
        & \multicolumn{2}{c}{\textbf{Controllability}}
        & \multicolumn{2}{c}{\textbf{Scene Consistency}}
        & \multicolumn{2}{c}{\textbf{Instance Consistency}} \\
        \cmidrule(lr){3-4}
        \cmidrule(lr){5-6}
        \cmidrule(lr){7-8}

        \textbf{Method}
        & \textbf{FVD} $\downarrow$
        & \textbf{ADE} $\downarrow$
        & \textbf{DTW} $\downarrow$
        & \textbf{CD} $\downarrow$
        & \textbf{RE} $\downarrow$
        & \textbf{Hit} $\uparrow$
        & \textbf{LE} $\downarrow$ \\
        \midrule
        
        HunyuanVideo-1.5
        & 1082.4 & 11.96 & 573 & 2.08 & \underline{0.61} & 0.322 & 10.92 \\
                
        Wan2.2-I2V
        & 444.2 & 22.95 & 1189 & \underline{1.82} & 0.64 & 0.302 & 11.16 \\
        
        MagicDrive-V2
        & 2815.4 & 10.71 & 546 & 19.53 & 1.27 & 0.007 & 12.00 \\

        ShareVerse
        & 228.0 & 15.25 & 399 & 4.21 & 1.11 & 0.159 & \underline{9.40} \\
        
        Cosmos3
        & \textbf{103.2} & \underline{5.69} & \underline{241} & 3.62 & 1.27 & \underline{0.373} & 9.83 \\

        \midrule
        \textbf{CoDrive}
        & \underline{108.8}
        & \textbf{3.05}
        & \textbf{116}
        & \textbf{1.64}
        & \textbf{0.52}
        & \textbf{0.482}
        & \textbf{5.81} \\

        \bottomrule
    \end{tabular}%
    }
\end{table*}

Table~\ref{tab:quantitative-comparison} shows that CoDrive performs favorably across trajectory controllability, scene consistency, and instance consistency, while maintaining competitive FVD.
Among pose-controllable baselines, CoDrive obtains the lowest ADE and DTW.
Compared with independently composed or existing multi-agent generation baselines, it also achieves better cross-agent consistency.
For trajectory control, CoDrive achieves the lowest ADE (3.05) and DTW (116) among the evaluated methods.
For scene consistency, CoDrive obtains the lowest CD (1.64) and RE (0.52), indicating better cross-agent scene agreement under the commanded camera geometry.
For instance consistency, CoDrive achieves the highest Hit Rate (0.482) and the lowest LE (5.81), indicating more accurate cross-agent rendering and localization of the participating vehicles.
Meanwhile, its FVD of 108.8 remains comparable to the best baseline result of 103.2, suggesting that the gains in cross-agent consistency do not require a substantial degradation in perceptual video quality.

\subsection{Ablation Studies}

\begin{table*}[t]
    \centering
    \caption{\textbf{Component ablation.} Component ablation on CoDrive-Bench. We isolate Interleaved Global-Local Self-Attention (IGLA) and Global Camera Geometry Injection (GCGI) respectively and verify their complementary effects.}
    \label{tab:component-ablation}
    \resizebox{0.8\textwidth}{!}{%
    \begin{tabular}{ccccccccc}
        \toprule
        & &
        & \multicolumn{2}{c}{\textbf{Controllability}}
        & \multicolumn{2}{c}{\textbf{Scene Consistency}}
        & \multicolumn{2}{c}{\textbf{Instance Consistency}} \\
        \cmidrule(lr){4-5}
        \cmidrule(lr){6-7}
        \cmidrule(lr){8-9}

        IGLA
        & GCGI
        & \textbf{FVD} $\downarrow$
        & \textbf{ADE} $\downarrow$
        & \textbf{DTW} $\downarrow$
        & \textbf{CD} $\downarrow$
        & \textbf{RE} $\downarrow$
        & \textbf{Hit} $\uparrow$
        & \textbf{LE} $\downarrow$ \\
        \midrule
        
        
        & $\checkmark$ 
        & 161.2 & 6.71 & 321 & 1.77 & 0.59 & 0.409 & 7.25 \\
                
        $\checkmark$ & 
        & 238.1 & 13.40 & 688 & 1.75 & 0.55 & 0.342 & 8.28 \\


        $\checkmark$ & $\checkmark$
        & \textbf{108.8}
        & \textbf{3.05}
        & \textbf{116}
        & \textbf{1.64}
        & \textbf{0.52}
        & \textbf{0.482}
        & \textbf{5.81} \\

        \bottomrule
    \end{tabular}%
    }
\end{table*}

\begin{table*}[t]
    \centering
    \caption{\textbf{Impact of mixed-task fine-tuning on performance in real-world scenarios.}}
    \label{tab:training-ablation}
    \resizebox{0.85\textwidth}{!}{%
    \begin{tabular}{cccccccc}
        \toprule
        &
        & \multicolumn{2}{c}{\textbf{Controllability}}
        & \multicolumn{2}{c}{\textbf{Scene Consistency}}
        & \multicolumn{2}{c}{\textbf{Instance Consistency}} \\
        \cmidrule(lr){3-4}
        \cmidrule(lr){5-6}
        \cmidrule(lr){7-8}

        & \textbf{FVD} $\downarrow$
        & \textbf{ADE} $\downarrow$
        & \textbf{DTW} $\downarrow$
        & \textbf{CD} $\downarrow$
        & \textbf{RE} $\downarrow$
        & \textbf{Hit} $\uparrow$
        & \textbf{LE} $\downarrow$ \\
        \midrule
                 
        w/o Mixed Fine-tuning
        & 282.9 & 6.86 & 343 & 1.68 & 0.47 & 0.560 & 7.69 \\

        w/ Mixed Fine-tuning
        & \textbf{104.3}
        & \textbf{1.53}
        & \textbf{61}
        & \textbf{1.57}
        & \textbf{0.45}
        & \textbf{0.646}
        & \textbf{5.99} \\

        \bottomrule
    \end{tabular}%
    }
\end{table*}

\paragraph{Interleaved Global-Local Self-Attention.}
As shown in Table~\ref{tab:component-ablation}, removing IGLA degrades performance across trajectory, scene-consistency, instance-consistency, and visual-quality metrics.
Without IGLA, the visual realism of the generated videos decreases, as evidenced by the FVD worsening from 108.8 in the full model to 161.2.
Furthermore, cross-agent scene consistency is compromised, with the CD increasing from 1.64 to 1.77 and the RE metric rising from 0.52 to 0.59.
The omission of IGLA also negatively impacts instance consistency, causing the Soft Hit Rate to drop from 0.482 to 0.409 and LE to increase from 5.81 to 7.25.
Controllability also suffers, as ADE increases to 6.71 and DTW reaches 321.
These results indicate that interleaving local and global attention contributes to both cross-agent consistency and trajectory controllability.

\paragraph{Global Camera Geometry Injection.}
Table~\ref{tab:component-ablation} also shows the contribution of Global Camera Geometry Injection (GCGI).
Without GCGI, trajectory controllability degrades substantially.
This is highlighted by a sharp increase in ADE to 13.40 and DTW to 688, compared to the full model's 3.05 and 116 respectively.
Additionally, removing GCGI harms overall video quality, pushing the FVD up to 238.1.
Cross-agent instance consistency is also heavily degraded without this explicit geometric guidance, leading to the lowest Soft Hit Rate of 0.342 and a high LE of 8.28.
These results suggest that explicit camera-geometry conditioning complements cross-agent attention, particularly for trajectory control and spatial alignment.

\paragraph{Mixed-Task Fine-Tuning.}
As shown in Table~\ref{tab:training-ablation}, mixed-task fine-tuning consistently improves performance on the real-world split across the reported metrics.
It reduces FVD from 282.9 to 104.3, ADE from 6.86 to 1.53, and DTW from 343 to 61, while increasing Hit Rate from 0.560 to 0.646 and reducing LE from 7.69 to 5.99.
These results suggest that incorporating real-world single-agent data during mixed-task fine-tuning improves transfer to real-world scenarios while retaining the cross-agent supervision learned from simulated dual-agent data.

\subsection{Qualitative Results}

\begin{figure*}[t!]
    \centering
        \includegraphics[width=0.95\linewidth]{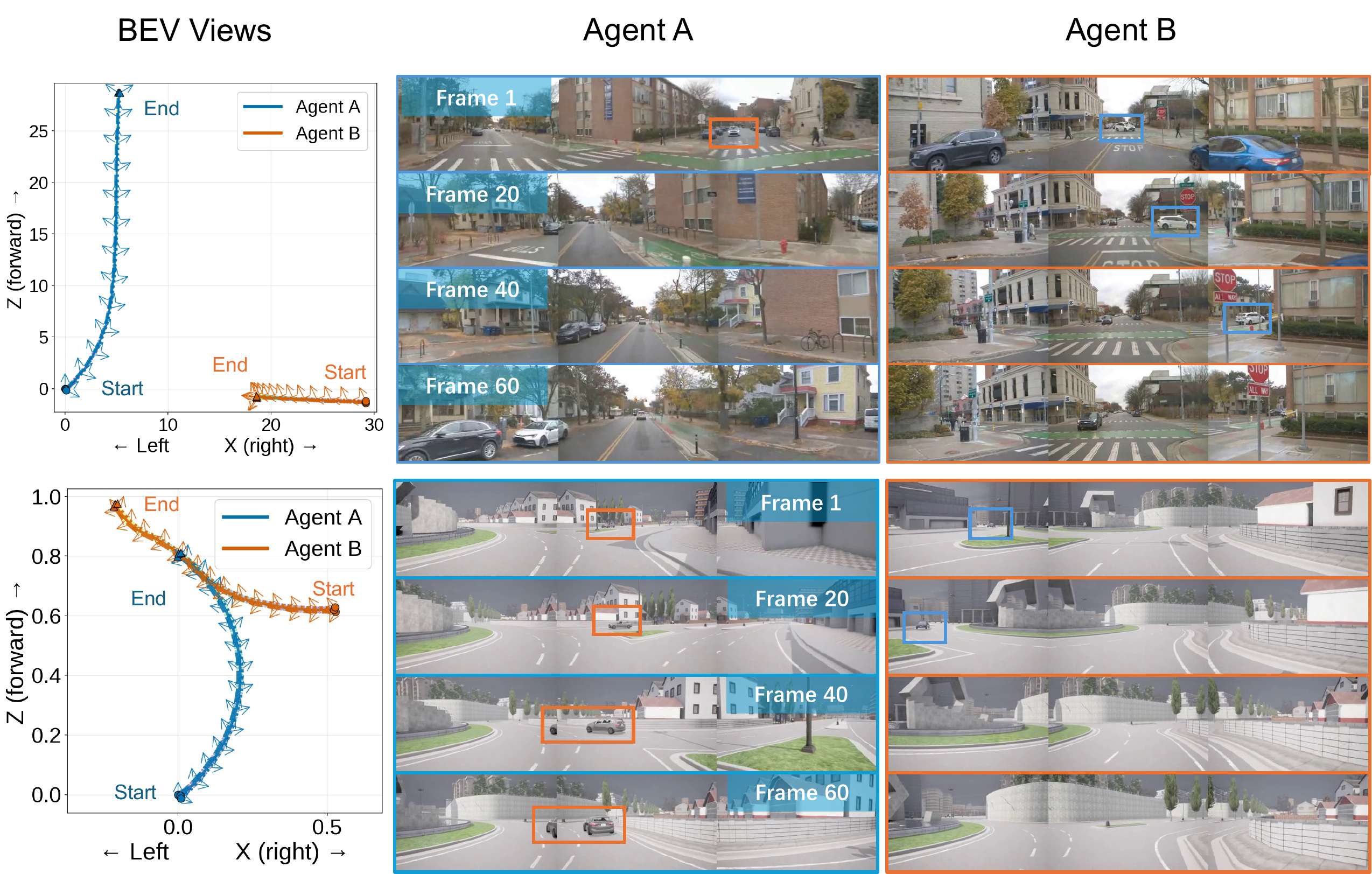}
    \caption{\textbf{Qualitative results on both real-world and synthetic data.} Blue image borders indicate Agent A’s three views, while orange image borders indicate Agent B’s three views. Blue bounding boxes mark Agent A in Agent B’s views, whereas orange bounding boxes mark Agent B in Agent A’s views.}
\label{fig:qualitative-results}
\end{figure*}

Figure~\ref{fig:qualitative-results} presents qualitative results in both real-world and synthetic environments.
CoDrive follows the trajectories while maintaining coherent scene layouts across views.
Shared static structures remain visually and geometrically compatible between agents, while the participating vehicles appear at positions and scales consistent with their relative poses and maintain continuous motion across frames.

%% file: sections/6-conclusion.tex
\section{Conclusion and Future Work}

We present CoDrive, a framework for jointly generating multi-view observations of vehicles that share the same dynamic driving scene.
We also introduce CoDrive-Bench to evaluate camera controllability, cross-agent scene consistency, and instance consistency.
Experiments on CoDrive-Bench show that CoDrive improves cross-agent consistency and trajectory controllability while maintaining competitive visual quality.
Our current implementation focuses on dual-vehicle interactions. Extending shared-world video generation to larger numbers of interacting vehicles and incorporating real-world multi-vehicle training data are important directions for future work.

%% file: sections/7-appendix.tex
\section{Method Details}

\subsection{Details of Synthetic Data Collection}\label{sec:appendix-data}

\paragraph{Overview.}
We construct a CARLA-based synthetic data collection pipeline for generating synchronized dual-vehicle, multi-view driving interactions.
Each synthetic sample contains two focal vehicles, denoted as Agent A and Agent B, following a controlled pair of interacting trajectories.
Three forward-facing RGB cameras are mounted on each vehicle, resulting in six temporally synchronized video streams per sample.
Along with the RGB observations, we record frame-wise vehicle poses, camera extrinsics and intrinsics, and structured metadata describing the environment and interaction.
After quality filtering, we retain 15K two-vehicle interaction clips for training.

\paragraph{Environment configurations.}
To increase visual and geometric diversity, we sample simulation environments from eight CARLA maps  containing diverse road layouts, including urban intersections, multi-lane roads, residential streets, curved roads, and complex junctions.

For each sample, we additionally select one of eleven CARLA weather presets:
\texttt{ClearNoon}, \texttt{CloudyNoon}, \texttt{WetNoon},
\texttt{WetCloudyNoon}, \texttt{SoftRainNoon},
\texttt{MidRainyNoon}, \texttt{HardRainNoon},
\texttt{ClearSunset}, \texttt{CloudySunset},
\texttt{WetSunset}, and \texttt{SoftRainSunset}.
Maps, weather conditions, and target interaction types are sampled uniformly.
All random variables are generated from case-specific deterministic seeds, making the collection process reproducible while maintaining diversity across samples.

\paragraph{Interaction types.}
We consider six vehicle interaction modes:
(i) intersection crossing,
(ii) same-direction following,
(iii) oncoming passing,
(iv) turning interactions,
(v) merging or cut-in, and
(vi) parallel-lane driving.
Their semantic definitions are summarized in Table~\ref{tab:carla-interactions}.

\begin{table*}[h]
\centering
\small
\caption{Vehicle interaction modes used in the synthetic data collection pipeline.}
\label{tab:carla-interactions}
\begin{tabular}{p{0.23\textwidth} p{0.69\textwidth}}
\toprule
Interaction mode & Description \\
\midrule
Intersection crossing &
The two vehicles approach a common region from substantially different directions and interact near an intersection or road crossing. \\

Same-direction following &
The two vehicles travel in approximately the same direction while maintaining a relatively short longitudinal distance. \\

Oncoming passing &
The two vehicles approach one another with nearly opposite headings and pass each other on opposing lanes or nearby roads. \\

Turning interaction &
At least one of the two vehicles undergoes a significant heading change while the vehicles remain spatially close. \\

Merging or cut-in &
The vehicles have approximately aligned headings and become substantially closer over time, covering road merging, lane convergence, and cut-in-like interactions. \\

Parallel-lane driving &
The two vehicles travel on approximately parallel roads or lanes without satisfying the stronger geometric conditions of the other interaction modes. \\
\bottomrule
\end{tabular}
\end{table*}

\paragraph{Trajectory generation.}
For each sample, we first randomly select two valid spawn points from the CARLA road network.
The speed of each focal vehicle is independently sampled from a predefined range, which is $4.8$--$8.8\,\mathrm{m/s}$ in our implementation.
Starting from the selected spawn points, we construct trajectories by iteratively querying successor waypoints from the CARLA road graph.
The distance between consecutive waypoints is determined by the sampled speed and simulation frame rate.
At road junctions, alternative waypoint branches are randomly selected, with a preference for turning branches when generating crossing, turning, or merging interactions.

Trajectory diversity is introduced by varying initial locations, vehicle speeds, road branches, and turning choices.
Vehicle categories and colors are also randomized independently for the two focal agents.
Agent A is sampled from a warm-colored palette and agent B from a complementary cool-colored palette, making the two focal vehicles easier to distinguish across views.

For each target interaction type, we generate multiple trajectory-pair candidates and evaluate their geometric relationships.
Let $\mathbf{p}_A^t$ and $\mathbf{p}_B^t$ denote the positions of the two vehicles at frame $t$, and let $\theta_A^t$ and $\theta_B^t$ denote their headings.
We compute the inter-agent distance
\begin{equation}
    d_t = \left\|\mathbf{p}_A^t-\mathbf{p}_B^t\right\|_2,
\end{equation}
as well as their wrapped heading difference
\begin{equation}
    \Delta\theta_t =
    \left|
    \operatorname{wrap}
    \left(\theta_A^t-\theta_B^t\right)
    \right|.
\end{equation}
The minimum distance, mean heading difference, heading difference at the closest frame, distance variation over time, and accumulated turning angles are used to classify and rank candidate interactions.
In particular, crossing interactions favor small minimum distances and large heading differences; following interactions favor small mean heading differences; oncoming interactions favor heading differences close to $180^\circ$; and merging interactions favor approximately aligned headings together with a substantial reduction in inter-agent distance.
Among the valid candidates, we select the trajectory pair that best matches the target interaction while maintaining sufficient spatial proximity and mutual visibility.

\paragraph{Kinematic trajectory replay.}
After selecting a valid trajectory pair, the two focal vehicles are replayed kinematically in the simulator.
Specifically, vehicle physics are disabled for the two focal agents, and their transforms are updated to the corresponding trajectory waypoints before each simulation tick.
This design avoids trajectory deviations caused by stochastic low-level vehicle control and ensures that all intended interactions are reproduced consistently.
Background actors, in contrast, can be controlled by CARLA's Traffic Manager and therefore retain independent motion dynamics.

\paragraph{Background distractors.}
The pipeline supports the insertion of task-irrelevant background vehicles and pedestrians.
Background vehicle types, colors, spawn locations, and driving speeds are randomized.
They are controlled by a synchronized CARLA Traffic Manager and are prevented from spawning within a protected corridor around the focal trajectories.
To ensure that distractors are visible without excessively interfering with the target interaction, the sampler can bias a configurable fraction of background actors toward spawn points near the focal routes.
Optional pedestrians are sampled from navigable regions using a similar distance-based protection strategy.
We reject a sample if background density constraints are enabled but the required number of visible or nearby distractors cannot be generated.

Each focal vehicle is additionally equipped with a collision sensor.
A sample is discarded if either focal vehicle triggers a collision event during collection, preventing invalid examples caused by overlapping actors or unintended collisions with background traffic.

\paragraph{Multi-view camera configuration.}
Each focal vehicle carries a three-camera rig consisting of left, center, and right forward-facing RGB cameras.
In the 81-frame configuration used in our experiments, the left, center, and right cameras have relative yaw angles of $-55^\circ$, $0^\circ$, and $55^\circ$, respectively.
All cameras use a pitch angle of $-4^\circ$ and are mounted approximately $1.65\,\mathrm{m}$ above the vehicle reference point.
The lateral offsets of the left and right cameras are approximately $-0.35\,\mathrm{m}$ and $0.35\,\mathrm{m}$, respectively.
The center camera is mounted slightly farther forward than the side cameras.

We record the following six RGB streams:
\begin{equation}
\begin{split}
\mathcal{V} = \{&
A_{\mathrm{left}}, A_{\mathrm{center}}, A_{\mathrm{right}},\\
&
B_{\mathrm{left}}, B_{\mathrm{center}}, B_{\mathrm{right}}
\}.
\end{split}
\end{equation}
The ordering remains fixed for all samples.
The configuration used for our dataset contains 81 frames captured at 10 frames per second, corresponding to approximately eight seconds of simulated interaction.
The images are rendered at $640\times384$ resolution with a horizontal field of view of $70^\circ$.
For each camera, the intrinsic calibration matrix is computed from the image resolution and field of view, while its world-coordinate extrinsic pose is recorded at every frame.

\paragraph{Temporal synchronization.}
CARLA is operated in synchronous mode with a fixed simulation interval of $0.1$ seconds.
Before recording, we perform several warm-up simulation ticks to initialize the rendering and sensor pipelines, after which all pending sensor messages are discarded.
At each recording step, the transforms of both focal vehicles are first updated, followed by exactly one call to the simulator tick function.
The six RGB observations and auxiliary sensor measurements are then retrieved according to the returned CARLA frame identifier.
This procedure ensures that all views in a multi-view frame correspond to the same simulation state.

\paragraph{Quality filtering.}
We apply geometric and visibility constraints before accepting a trajectory pair.
First, the minimum distance between the focal vehicles is constrained to be between $8\,\mathrm{m}$ and $34\,\mathrm{m}$, preventing both physically implausible overlap and interactions that are too distant.
The vehicles must remain within $30\,\mathrm{m}$ of each other for at least 24 frames and within $22\,\mathrm{m}$ for at least 10 frames.

We further perform a camera-frustum-based visibility test.
At each frame, the relative position of one vehicle is projected into the viewing sectors of the three cameras mounted on the other vehicle.
A vehicle is considered geometrically visible if it is in front of at least one camera, lies within its horizontal field of view, and is no farther than $42\,\mathrm{m}$ away.
We require at least one agent to observe the other for a minimum of 24 frames.
In addition, each agent must observe its counterpart for at least eight frames.
Although this geometric test does not explicitly model object-level occlusions, it efficiently removes trajectory pairs in which the two focal agents rarely appear in one another's camera views.

If route generation, visibility validation, collision checking, sensor acquisition, or video encoding fails, the current sample is discarded and regenerated using a different case-specific random seed.
The collection configuration allows multiple attempts per sample, which substantially improves the yield of valid interactions without relaxing the quality criteria.

\paragraph{Auxiliary geometry and metadata.}
For every frame, we store the world-coordinate poses of agents A and B, their relative longitudinal and lateral relationships, and the world-coordinate extrinsics of all six cameras.
We additionally record the states of visible scene actors, camera intrinsics, camera rig parameters, map identity, weather condition, interaction category, and collection quality statistics.
A structured textual description is generated from the scene configuration and interaction metadata.
The six independent RGB videos, rather than a spatially tiled preview, are used as the multi-view training data.

Finally, we scan all successfully completed cases and retain 15K clips that satisfy the trajectory, visibility, collision, and sensor-integrity requirements.
Each retained sample consists of six synchronized RGB videos together with the corresponding agent and camera poses, providing controlled yet diverse dual-agent supervision for model training.

\subsection{PRoPE Parameterization and Gauge Invariance}
\label{sec:appendix-prope}

\paragraph{Coordinate convention.}
Let
$\mathbf E_a={}^{C_a}\mathbf T_W$ be the world-to-camera transform and
$\mathbf C_a=\mathbf E_a^{-1}={}^{W}\mathbf T_{C_a}$ its inverse.
The implementation first computes
\begin{equation}
  \bar{\mathbf C}_a
  =
  \mathbf C_{a_0}^{-1}\mathbf C_a
  =
  {}^{C_{a_0}}\mathbf T_{C_a},
\end{equation}
and then returns
\begin{equation}
  \widetilde{\mathbf E}_a
  =
  \bar{\mathbf C}_a^{-1}
  =
  \mathbf E_a\mathbf E_{a_0}^{-1}
  =
  {}^{C_a}\mathbf T_{C_{a_0}}.
\end{equation}
Thus, $\widetilde{\mathbf E}_a$ expresses every camera in the coordinate frame of the same reference camera $a_0$ , providing a shared gauge across agents and views.

\paragraph{Tensor parameterization.}
For each batch element, attention head, and image token, the learned
features have shape
\begin{equation}
  \mathbf Q,\mathbf K,\mathbf V
  \in\mathbb R^{B\times N_h\times L\times d_h},
  \qquad d_h=4R.
\end{equation}
The wide latent is flattened in $(t,h,v,u)$ order, with
\begin{equation}
  L=T_pH_pVW_p,
  \qquad
  \ell=((tH_p+h)V+v)W_p+u.
\end{equation}
Thus, token $\ell$ receives the transform of camera $(t,v)$.
All spatial tokens belonging to the same view and latent time step therefore share the same camera transform.
Writing
$\mathbf q_\ell=[
(\mathbf q_\ell^{(1)})^\top,\ldots,
(\mathbf q_\ell^{(R)})^\top]^\top$
with $\mathbf q_\ell^{(r)}\in\mathbb R^4$, the implementation applies
$\mathbf P_\ell^\top$ independently to every query group and
$\mathbf P_\ell^{-1}$ independently to every key/value group.
This is equivalent to applying
\begin{equation}
  \mathcal P_\ell
  =
  \operatorname{diag}(
  \underbrace{\mathbf P_\ell,\ldots,\mathbf P_\ell}_{R\ \mathrm{times}})
  =
  \mathbf I_R\otimes\mathbf P_\ell.
\end{equation}
The construction requires $d_h$ to be divisible by four, with the same $4\times4$ projective transform independently applied to each channel group.

\paragraph{World-gauge invariance.}
Consider an arbitrary change of the source world coordinates
$\mathbf x_{W'}=\mathbf G\mathbf x_W$. The camera-to-world poses then become
$\mathbf C'_a=\mathbf G\mathbf C_a$. Their canonicalized versions satisfy
\begin{equation}
  \bar{\mathbf C}'_a
  =
  (\mathbf G\mathbf C_{a_0})^{-1}
  (\mathbf G\mathbf C_a)
  =
  \mathbf C_{a_0}^{-1}
  \mathbf C_a
  =
  \bar{\mathbf C}_a.
\end{equation}
Consequently,
$\widetilde{\mathbf E}'_a=\widetilde{\mathbf E}_a$,
$\mathbf P'_a=\mathbf P_a$, and
$\mathcal P'_a=\mathcal P_a$.
Therefore all transformed Q/K/V features, attention logits, and transported
outputs are unchanged by an arbitrary reparameterization of the original
world frame.
In other words, PRoPE depends on relative camera geometry rather than on the arbitrary coordinate system used to represent the scene.

\paragraph{Reference cancellation.}
The pairwise projective operator is
\begin{align}
  \mathbf P_i\mathbf P_j^{-1}
  &=
  \widehat{\mathbf K}_i
  \mathbf E_i\mathbf E_{a_0}^{-1}
  \left(
    \widehat{\mathbf K}_j
    \mathbf E_j\mathbf E_{a_0}^{-1}
  \right)^{-1}\\
  &=
  \widehat{\mathbf K}_i
  \mathbf E_i\mathbf E_{a_0}^{-1}
  \mathbf E_{a_0}\mathbf E_j^{-1}
  \widehat{\mathbf K}_j^{-1}\\
  &=
  \widehat{\mathbf K}_i
  \mathbf E_i\mathbf E_j^{-1}
  \widehat{\mathbf K}_j^{-1}.
\end{align}
Hence, for image tokens $i$ and $j$,
\begin{equation}
  \left\langle
    \mathcal P_i^\top\mathbf q_i,
    \mathcal P_j^{-1}\mathbf k_j
  \right\rangle
  =
  \mathbf q_i^\top
  \left[
    \mathbf I_R\otimes
    \left(
      \widehat{\mathbf K}_i
      \mathbf E_i\mathbf E_j^{-1}
      \widehat{\mathbf K}_j^{-1}
    \right)
  \right]
  \mathbf k_j.
\end{equation}
The shared reference camera cancels exactly from the pairwise operator.
The extrinsic component $\mathbf E_i\mathbf E_j^{-1}$ maps camera-$j$ coordinates to camera-$i$ coordinates, while the lifted intrinsics calibrate this relative transform
in the PRoPE feature space.
Consequently, pairwise attention depends only on the calibrated relative relationship between cameras $i$ and $j$, not on which camera was selected as the canonical reference.

\paragraph{Output-side transport.}
For the image-to-image component, the PRoPE output at token $i$ is
\begin{equation}
  \mathbf o_i^{\mathrm{PRoPE}}
  =
  \mathcal P_i
  \sum_j
  \alpha_{ij}
  \mathcal P_j^{-1}\mathbf v_j,
  \qquad
  \alpha_{ij}
  =
  \operatorname{softmax}_j
  \left(
    \frac{
      \mathbf q_i^\top
      \mathcal P_i\mathcal P_j^{-1}
      \mathbf k_j
    }{\sqrt{d_h}}
  \right).
\end{equation}
Thus, both attention weights and value transport are governed by the same calibrated pairwise camera geometry.
The text Q/K/V features remain untransformed in the implemented auxiliary branch.

\subsection{Training Details}\label{sec:appendix-training}

\paragraph{Stage 1: Backbone domain adaptation.}
In this stage, we perform text-image-to-video (TI2V) training on the backbone using large-scale driving video data with text annotations. The objective is to adapt the backbone to the domain of driving video generation, thereby reducing the training burden in subsequent stages. The only data requirement at this stage is the availability of text annotations; additional information such as camera poses is not required. Therefore, the entirety of our dataset can be used for training. We segment front-view videos from public datasets, including OpenDV-2K~\citep{yang2024genad}, nuScenes~\citep{caesar2020nuscenes}, Waymo~\citep{sun2020waymoscalability}, and PandaSet~\citep{xiao2021pandaset}, as well as from our collected CARLA synthetic dataset. This process yields approximately 900K video clips, each containing 61–81 frames. We then use a specialized vision-language model to generate text annotations for these clips. The model is subsequently trained for 15K iterations with a batch size of 64 at dynamic resolution. The resulting checkpoint is used as the initialization for the next stage.

\paragraph{Stage 2: Single-vehicle, single-view camera controllability training.}
In this stage, we train the model to acquire fundamental camera trajectory control capabilities. From the Stage 1 training data, we select all videos annotated with ego-camera pose trajectories, resulting in approximately 45K video clips. We augment the attention layers of the Stage 1 checkpoint with zero-initialized PRoPE Attention branches to learn camera trajectory control. Training is conducted in two phases. First, the model is trained at a resolution of $192 \times 320$ for $15$K iterations with a batch size of 32. The resolution is then increased to $384 \times 640$, and the model is trained for another 15K iterations with a batch size of 16 and learning rate of $1 \times 10^{-5}$.

\paragraph{Stage 3: Single-vehicle, multi-view extension training.}
In this stage, we extend the camera-controllable single-view video generation model to single-vehicle, multi-view generation. Since all 45K video samples with camera pose trajectory annotations used in Stage 2 contain at least a shared set of three front-facing views, we continue to use the same dataset in Stage 3. However, instead of using only a single front-facing view, we jointly use all three views for training. The model is first trained at a resolution of $192 \times 960$ for 15K iterations. The resolution is then increased to $384 \times 1920$, followed by another 15K training iterations. The batch size is set to 16 for both phases. The learning rate of this stage is $1 \times 10^{-5}$.

\paragraph{Stage 4: Multi-vehicle, multi-view consistency training.}
In this stage, we train the model to generate videos with cross-agent consistency using cross-agent, multi-view data. Since real-world multi-agent video data is extremely scarce, we use only the 15K dual-agent interaction samples collected from the CARLA simulation environment for training. During this stage, we apply the Interleaved Global-Local Self-Attention and Global Camera Geometry Injection proposed in the main paper during the model’s forward pass, enabling the model to efficiently learn cross-agent spatial consistency. At this stage, the model is trained to generate 61-frame videos at a resolution of $384 \times 960$, with a batch size of 16 and learning rate of $3 \times 10^{-5}$, for 10K iterations.

\paragraph{Stage 5: Mixed-task fine-tuning.}
Since our dual-agent training data consists exclusively of data from simulated environments, we introduce an additional stage of mixed-task fine-tuning. During this stage, the model randomly alternates between training on real-world single-vehicle, multi-view data and simulated multi-vehicle, multi-view data. The objective is to combine the visual realism learned from real-world data with the cross-vehicle consistency learned from simulated data. The two training tasks are sampled with equal probability, at 50\% each. The batch size is set to 16, and the learning rate is set to $1 \times 10^{-5}$.

\section{Benchmark Details}\label{sec:appendix-bench}

\subsection{Metrics Definition}\label{sec:appendix-bench-metrics}

\subsubsection{Controllability}

Following DrivingGen~\citep{zhou2026drivinggen}, we estimate monocular depth for each generated video and use the resulting depth together with visual SLAM to recover the camera trajectory.
We use the same depth-estimation and visual-SLAM pipelines as DrivingGen.

Because monocular trajectory recovery determines the camera path only up to an arbitrary rigid gauge, the recovered and commanded trajectories are brought into a common frame before either metric is evaluated.
We express the ground-truth trajectory in its own ego frame, placing its first position at the origin and aligning its initial heading with the $+x$ axis, and then rotate the recovered trajectory onto it by solving an orthogonal Procrustes problem with the first frame pinned to the origin.
Both trajectories are finally smoothed with a Savitzky--Golay filter, since frame-to-frame recovery is noisy.
We do not optimize a scale factor during alignment, so errors in traveled distance remain part of the controllability metric.
Accordingly, we report Average Displacement Error rather than the SLAM convention of Absolute Trajectory Error after full similarity alignment.

\paragraph{Average Displacement Error (ADE).} ADE is the average Euclidean distance between the corresponding positions at each time step.
For a predicted trajectory \(p_t\) of length \(T\) and its ground-truth trajectory \(g_t\), both expressed in the aligned frame described above:
\[
\mathrm{ADE}(p,g)
=
\frac{1}{T}
\sum_{t=1}^{T}
\left\|p_t-g_t\right\|_2.
\]

ADE is measured in meters, and a lower value indicates better performance. It quantifies the deviation between the trajectory of the model-generated video and the given input trajectory. Each two-vehicle clip contributes two independent single-agent samples, one per vehicle, since each vehicle is driven by its own commanded trajectory.

\paragraph{Dynamic Time Warping (DTW).} 
We employ the classical Dynamic Time Warping (DTW) algorithm, where the pairwise cost is defined as the two-dimensional Euclidean distance:
\[
d(i,j)=\left\|p_i-g_j\right\|_2.
\]

The dynamic programming recurrence is given by:
\[
D(i,j)
=
d(i,j)
+
\min
\left\{
\begin{array}{l}
D(i-1,j),\\
D(i,j-1),\\
D(i-1,j-1)
\end{array}
\right..
\]

The boundary condition is:
\[
D(0,0)=0.
\]

All other entries in the zeroth row and zeroth column are initialized to infinity. The final result is:
\[
\mathrm{DTW}(p,g)=D(T_p,T_g).
\]

\subsubsection{Scene Consistency}

A two-vehicle scenario consists of two agents (vehicles), $a$ and $b$, each equipped with a set of forward-facing cameras (views). Let $\mathcal{V}_a$ and $\mathcal{V}_b$ denote the sets of views associated with the two agents.
For any frame and any view $c$, the following three quantities are known or can be estimated:
\begin{itemize}
    \item A metric depth map $D_c \in \mathbb{R}^{H \times W}$, which provides the per-pixel metric depth in meters estimated by a monocular depth estimator;
    
    \item Camera intrinsics $K_c$, including the focal lengths $(f_x,f_y)$ and the principal point $(c_x,c_y)$;
    
    \item Camera extrinsics $T_c \in SE(3)$, representing the rigid-body transformation from the world coordinate system to the camera coordinate system. Its inverse, $T_c^{-1}$, maps points from the camera coordinate system to the world coordinate system.
\end{itemize}

Define the back-projection operator $\Phi_c$, which maps a pixel $(u,v)$ in view $c$, together with its depth $z = D_c(u,v)$, to a three-dimensional point in the world coordinate system:
$$
    \Phi_c(u,v)
    =
    T_c^{-1}
    \begin{bmatrix}
        \dfrac{u-c_x}{f_x} z \\
        \dfrac{v-c_y}{f_y} z \\
        z
    \end{bmatrix}.
$$

Correspondingly, the projection operator $\pi_c(X)$ projects a world point $X$ onto view $c$ and returns its pixel coordinates and depth in the camera coordinate system. A world point $X$ is said to be observed by agent $g$ if there exists a view $c \in \mathcal{V}_g$ such that $\pi_c(X)$ lies within the image boundaries and its camera-frame depth belongs to the interval
$$
    (0,d_{\max}].
$$

Both metrics are computed \textbf{after foreground removal}. Pixels corresponding to dynamic vehicles are removed using vehicle detection bounding boxes, leaving only the set of static-background pixels
$$
    \mathcal{S}_c
    \subseteq
    \{1,\ldots,W\} \times \{1,\ldots,H\}
$$
for view $c$.
The resulting metrics therefore focus on cross-agent agreement of static scene geometry rather than independently moving vehicles.

\paragraph{Chamfer Distance (CD).}
First, we construct the point clouds. The static-background point cloud of agent $a$ is defined as the union of the back-projected static pixels from all of its views:
\begin{equation}
    P_a
    =
    \left\{
        \Phi_c(u,v)
        \;\middle|\;
        c \in \mathcal{V}_a,\,
        (u,v) \in \mathcal{S}_c
    \right\},
    \qquad
    P_b \text{ is defined analogously.}
\end{equation}

We define the co-visible regions by retaining only the points that can be observed by the other agent:
\begin{equation}
    P_a^{\mathrm{cov}}
    =
    \left\{
        X \in P_a
        \;\middle|\;
        X \text{ is observed by } b
    \right\},
    \qquad
    P_b^{\mathrm{cov}}
    =
    \left\{
        X \in P_b
        \;\middle|\;
        X \text{ is observed by } a
    \right\}.
\end{equation}
Both point clouds are reconstructed directly in the commanded world frame using the camera extrinsics supplied to the generator.
CD is therefore a conditioned scene-consistency metric. It evaluates whether the generated observations are mutually compatible under the prescribed camera geometry, rather than isolating pose-invariant reconstruction quality.
Consequently, CD penalizes both mutually inconsistent scene reconstructions and inconsistencies caused by deviations from the supplied camera trajectories.
This coupling is intentional for our end-to-end conditioned generation setting, in which the observations are required to be consistent with both one another and the prescribed camera geometry.
The symmetric Chamfer distance is
\begin{equation}
    \mathrm{CD}
    =
    \frac{1}{2}
    \Bigg[
        \frac{1}{\left|P_a^{\mathrm{cov}}\right|}
        \sum_{p \in P_a^{\mathrm{cov}}}
        \min_{q \in P_b^{\mathrm{cov}}}
        \lVert p-q \rVert_2
        +
        \frac{1}{\left|P_b^{\mathrm{cov}}\right|}
        \sum_{q \in P_b^{\mathrm{cov}}}
        \min_{p \in P_a^{\mathrm{cov}}}
        \lVert q-p \rVert_2
    \Bigg].
\end{equation}

The metric is computed frame by frame and averaged over frames with sufficient co-visible regions. It is measured in meters, and a smaller value indicates greater consistency between the three-dimensional world reconstructions produced by the two agents.

\paragraph{Reprojection Depth Error (RE).}
RE measures cross-agent depth consistency through image-space reprojection under the supplied camera geometry.
Consider a cross-agent view pair $(c_A,c_B)$, where $c_A \in \mathcal{V}_a$ is the target view and $c_B \in \mathcal{V}_b$ is the source view.

We first perform cross-view reprojection. For each static-background pixel $(u,v)$ in the target view with depth $z = D_{c_A}(u,v)$, we first back-project it into the world coordinate system,
\begin{equation}
    X = \Phi_{c_A}(u,v),
\end{equation}
and then project it into the source view:
\begin{equation}
    \left( (u',v'),\, z_{\mathrm{pred}} \right)
    =
    \pi_{c_B}(X),
\end{equation}
where $z_{\mathrm{pred}}$ denotes the predicted depth of the point in the source camera coordinate system. The observed depth is obtained by bilinearly sampling the source depth map at $(u',v')$:
\begin{equation}
    z_{\mathrm{obs}}
    =
    D_{c_B}(u',v').
\end{equation}
The set of valid pixels, denoted by $\mathcal{P}$, consists of those satisfying all of the following conditions:
\begin{itemize}
    \item the reprojected pixel $(u',v')$ lies within the source image boundaries;
    \item both $z$ and $z_{\mathrm{pred}}$ fall within the valid depth range $(0,d_{\max}]$;
    \item both $(u,v)$ and $(u',v')$ belong to the static-background regions.
\end{itemize}
We compare depths in metric units, so disagreement in absolute scene scale contributes directly to the error.
The relative reprojection depth error is defined as
\begin{equation}
    E
    =
    \frac{1}{|\mathcal{P}|}
    \sum_{(u,v)\in\mathcal{P}}
    \frac{\left| z_{\mathrm{pred}} - z_{\mathrm{obs}} \right|}
         {z_{\mathrm{pred}}}.
\end{equation}

The final RE metric is obtained by averaging $E$ over all cross-agent view pairs and all frames.
A smaller value indicates better depth consistency between the two views on the image plane.

\subsubsection{Instance Consistency}

Each agent corresponds to one of the two focal vehicles, whose pose is determined from the camera extrinsics provided as generation conditions.
We take its reference position $X^\star$ to be the centroid of the agent's camera centers, $X^\star = \frac{1}{|\mathcal{V}_g|}\sum_{c \in \mathcal{V}_g} T_c^{-1}\mathbf{0}$, and its heading to be the optical axis of its front camera projected onto the horizontal plane.
We then represent the target as a three-dimensional bounding box centered at $X^\star$ and oriented by that heading.
The rig centroid is a proxy for the physical vehicle center. Because the same proxy is used for both the generated and reference geometry, the corresponding offset does not introduce a relative bias between them.

Given an observation direction---that is, a view $c$ of the observer agent, with the other agent treated as the target---we project $\mathcal{B}^\star$ onto view $c$ to obtain:
\begin{itemize}
    \item a predicted two-dimensional bounding box $b_{\mathrm{pred}}$, whose center is denoted by $m_{\mathrm{pred}}$ and whose diagonal length is denoted by $\ell$;
    
    \item a predicted distance $z_{\mathrm{pred}}$, defined as the depth of the target in the coordinate system of camera $c$.
\end{itemize}

For each generated view, we apply a vehicle detector to obtain frame-level bounding boxes and a multi-object tracker to associate them across time. Soft Hit Rate is evaluated on the resulting tracks, whereas LE is evaluated on individual matched detections.

We first define a common set of geometrically valid observations, $\Omega$, which is used by both instance-consistency metrics.
\[
    (\text{frame},\ \text{observation view } c,\ \text{other agent})
\]
that satisfy the following conditions: $\mathcal{B}^\star$ is visible in view $c$, lies entirely within the image without touching its boundaries, and satisfies
\begin{equation}
    z_{\mathrm{pred}} \leq d_{\max}.
\end{equation}
Both observation directions, namely $a$ observing $b$ and $b$ observing $a$, are included.
Thus, $\Omega$ contains all observations for which the geometry predicts that the observer should see the other vehicle.

Observations in which the projected box touches the image boundary or the target exceeds the maximum distance are excluded.
A partially out-of-frame vehicle or a distant vehicle beyond the detector's effective resolution may be missed because of the detector's limitations rather than the quality of the generated view.

For conciseness, we define a truncated linear decay kernel. Given thresholds $\tau < \kappa$,
\begin{equation}
    R(x;\tau,\kappa)
    =
    \begin{cases}
        1,
        & x \leq \tau, \\[2pt]
        \dfrac{\kappa-x}{\kappa-\tau},
        & \tau < x < \kappa, \\[4pt]
        0,
        & x \geq \kappa.
    \end{cases}
\end{equation}
The kernel decreases monotonically from $1$ to $0$.

\paragraph{Soft Hit Rate (Hit).}
This metric asks whether a \emph{persistent} vehicle track \emph{follows} the geometrically predicted trajectory throughout the co-visible window.
Since the predicted box sequence varies with the target vehicle’s motion, incorrect speed or heading causes the detected track to drift from the geometric prediction over time. The metric therefore evaluates temporal cross-agent consistency rather than frame-wise spatial coincidence alone.

Fix an observer view $c$ and let the other agent be the target. Let
\begin{equation}
    \mathcal{F}_c
    =
    \left\{
        t : (t,c,\text{target}) \in \Omega
    \right\},
    \qquad
    n_c = \left|\mathcal{F}_c\right|
\end{equation}
denote the frames of the co-visible window, i.e. those frames retained by the gating conditions above. For each $t \in \mathcal{F}_c$ the projection of $\mathcal{B}^\star$ supplies the predicted center $m_{\mathrm{pred}}(t)$, the diagonal length $\ell(t)$, and the predicted distance $z_{\mathrm{pred}}(t)$. Views with $n_c = 0$ are not scored, since the geometry never predicts a valid observation there.

We apply a multi-object tracker to view $c$, which assigns persistent identities across frames and yields a set of tracks $\mathcal{K}_c$. A track $k \in \mathcal{K}_c$ is observed on frames $\mathcal{T}_k$ with centers $m_k(t)$; detections that the tracker fails to associate carry no identity and are discarded, as they provide no evidence of persistence. Writing $\mathcal{C}_k = \mathcal{T}_k \cap \mathcal{F}_c$ for the frames shared with the predicted window, we define three quantities per track:
\begin{align}
    \mathrm{cov}_k
    &=
    \frac{\left|\mathcal{C}_k\right|}{n_c},
    \\[4pt]
    \delta_k
    &=
    \frac{1}{\left|\mathcal{C}_k\right|}
    \sum_{t \in \mathcal{C}_k}
    \frac{
        \left\lVert m_k(t)-m_{\mathrm{pred}}(t)\right\rVert_2
    }{\ell(t)},
    \\[4pt]
    g_k
    &=
    \frac{1}{\left|\mathcal{C}_k\right|}
    \sum_{t \in \mathcal{C}_k}
    R\!\left(
        \frac{
            \left|z_k(t)-z_{\mathrm{pred}}(t)\right|
        }{
            z_{\mathrm{pred}}(t)
        }\,
    \right),
\end{align}
namely the temporal \emph{coverage} of the predicted window, the mean scale-normalized \emph{trajectory deviation}, and the mean \emph{depth agreement}. Here $z_k(t)$ is the median depth inside the track's box at frame $t$. As before, each displacement is normalized by $\ell(t)$ so that near and distant targets are comparable, and the depth term uses a relative error because monocular depth error grows with distance. When depth is unavailable we set $g_k \equiv 1$, so depth gating can only tighten the criterion and never manufacture score.

Among the tracks that persist for a sufficient fraction of the window,
\begin{equation}
    \mathcal{K}_c^{\mathrm{adm}}
    =
    \left\{
        k \in \mathcal{K}_c :
        \mathrm{cov}_k \geq \rho
    \right\},
\end{equation}
we retain the one that best follows the prediction, $k^\star = \arg\min_{k \in \mathcal{K}_c^{\mathrm{adm}}} \delta_k$, and define the score of the observation direction as
\begin{equation}
    s_c
    =
    \begin{cases}
        \mathrm{cov}_{k^\star}
        \cdot
        R\!\left(\delta_{k^\star}\right)
        \cdot
        g_{k^\star},
        & \mathcal{K}_c^{\mathrm{adm}} \neq \emptyset, \\[4pt]
        0,
        & \text{otherwise.}
    \end{cases}
\end{equation}
The Soft Hit Rate is the mean of $s_c$ over all scored observation directions, covering both $a$ observing $b$ and $b$ observing $a$.

The score combines temporal coverage, image-space trajectory agreement, and relative-depth agreement, so each failure mode independently reduces the final score.
The coverage threshold $\rho$ further suppresses transient distractors whose tracks overlap the predicted path only briefly.

\paragraph{3D Localization Error (LE).}
The LE metric is computed only over the subset of observations $\mathcal{H}\subseteq\Omega$ that produce a binary hit, meaning that the overlap between the detected bounding box and the predicted bounding box exceeds a predefined threshold and a vehicle is successfully detected.
For each $\omega\in\mathcal{H}$, we take the center pixel of the matched detected bounding box and back-project it into the world coordinate system using the median depth within that box in the observation view. This yields the estimated three-dimensional position of the other vehicle from the observer's perspective:
\begin{equation}
    \hat{X}
    =
    \Phi_c\left(\text{center of the detected bounding box}\right).
\end{equation}

We then compare $\hat{X}$ with the geometrically specified ground-truth position $X^\star$:
\begin{equation}
    \mathrm{m\text{-}err}(\omega)
    =
    \left\lVert \hat{X}-X^\star \right\rVert_2,
    \qquad
    \mathrm{m\text{-}err}
    =
    \operatorname*{median}_{\omega\in\mathcal{H}}
    \mathrm{m\text{-}err}(\omega).
\end{equation}
We aggregate localization errors with the median because occasional large monocular-depth errors can otherwise dominate the mean.

LE is conditional on successful rendering and detection. It measures localization accuracy when the other vehicle is present, but not how often that vehicle is generated. Soft Hit Rate captures the complementary failure mode.
The two metrics should therefore be interpreted jointly. For example, low LE with low Hit indicates accurate localization when successful but frequent failures to render the target vehicle.

\begin{table*}[h]
    \centering
    \caption{\textbf{The scored observation set is constructed from pose-based criteria that apply identically to all methods, yielding an exactly paired comparison free of gating bias.}
    }
    \label{tab:perception-audit}
    \resizebox{0.7\textwidth}{!}{%
    \begin{tabular}{lcccc}
        \toprule
        & \multicolumn{2}{c}{\textbf{Synthetic}}
        & \multicolumn{2}{c}{\textbf{Real}} \\
        \cmidrule(lr){2-3}
        \cmidrule(lr){4-5}
        & \textbf{count} & \textbf{share} & \textbf{count} & \textbf{share} \\
        \midrule

        \multicolumn{5}{l}{\textit{Not resolvable by the camera rig}} \\
        \quad Target behind the observer     & 10{,}802 & 29.5\% & 13{,}255 & 36.2\% \\
        \quad Projects to ${<}0.3\%$ of image & 14{,}020 & 38.3\% & 14{,}606 & 39.9\% \\
        \midrule
        \textbf{Geometrically visible}       & \textbf{11{,}778} & \textbf{32.2\%} & \textbf{8{,}739} & \textbf{23.9\%} \\
        \midrule
        \multicolumn{5}{l}{\textit{Excluded by the metric, as a share of the visible set}} \\
        \quad Beyond $d_{\max}=35$\,m        &  2{,}676 & 22.7\% &     874 & 10.0\% \\
        \quad Box touches the frame edge     &  2{,}896 & 24.6\% &  2{,}257 & 25.8\% \\
        \midrule
        \textbf{Scored observations}         & \textbf{6{,}206} & \textbf{52.7\%} & \textbf{5{,}608} & \textbf{64.2\%} \\

        \bottomrule
    \end{tabular}%
    }
\end{table*}

\paragraph{Selection effect of instance metrics.}
LE is explicitly conditioned on successful rendering and detection, whereas Hit measures persistence and visibility. The scored observation set is constructed from transparent, pose-based criteria that apply identically to all methods, ensuring a fair and exactly paired comparison.
Table~\ref{tab:perception-audit} separates observations that no camera could resolve from the two conditions that reflect deliberate design choices, and reports the latter as a share of the geometrically visible set.
Observations where the target lies behind the observer or subtends too few pixels are excluded regardless of generation quality, because no camera could physically capture them.
Of the geometrically visible observations that remain, 52.7\% on CARLA and 64.2\% on real data survive the two deliberate exclusion criteria and are scored, yielding several thousand observations per domain.
Since every condition is a function solely of the ground-truth poses and never of the generated pixels, the same observations are scored for every method and for the recorded videos.
No method can be advantaged by the gating, and the comparison is exactly paired.

\subsection{Sensitivity Analysis of Consistency Metrics}

\subsubsection{Metric Results Variation under Controlled Perturbation}
\begin{figure*}[t!]
    \centering
        \includegraphics[width=\linewidth]{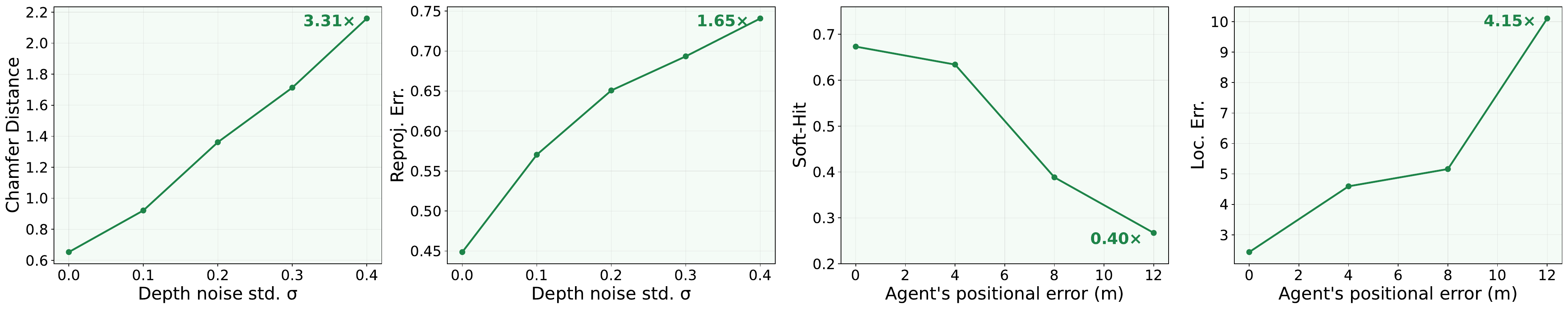}
    \caption{\textbf{Sensitivity analysis of CoDrive-Bench consistency metrics under controlled perturbations}. We progressively perturb the depth estimates and participating-agent position starting from the unperturbed reference setting. Increasing depth noise monotonically increases CD and RE, while increasing agent-position perturbation decreases Soft Hit Rate and increases LE. The annotations report the relative change between the largest perturbation and the unperturbed setting.}
\label{fig:metric-sens}
\end{figure*}

To examine whether the proposed consistency metrics respond to their intended failure modes, we conduct controlled perturbation experiments starting from the unperturbed reference video.
For scene consistency, we progressively inject Gaussian noise with standard deviation $\sigma$ into the depth maps and evaluate the resulting changes in CD and RE.
For instance consistency, we introduce increasing positional perturbations to the participating agent and measure the corresponding changes in Hit and LE.

As shown in Figure~\ref{fig:metric-sens}, all four metrics change monotonically with perturbation strength. Increasing the depth-noise standard deviation from 0 to 0.4 increases CD from approximately 0.65 to 2.15 (3.31×) and RE from 0.45 to 0.74 (1.65×). Similarly, increasing the agent position perturbation from 0 to 12m reduces Soft Hit Rate from approximately 0.67 to 0.27, while increasing LE from 2.4m to 10.1m. These monotonic trends provide a sanity check that the proposed metrics respond predictably to controlled violations of scene- and instance-level consistency.

\subsubsection{Metric Robustness to Evaluator Choice}

\begin{table*}[t]
    \centering
    \caption{\textbf{Robustness of CoDrive-Bench metrics to evaluator choice}. We compare model rankings obtained using alternative depth estimators for CD and RE, and alternative multi-object trackers for Hit and LE. Spearman’s $\rho$ is computed across all evaluated methods. Rankings remain highly consistent across evaluator variants, and CoDrive retains the top rank in all cases.}
    \label{tab:evaluator-robustness}
    \resizebox{0.8\textwidth}{!}{%
    \begin{tabular}{cccc}
        \toprule

        \textbf{Metric}
        & \textbf{Evaluator Variants}
        & \textbf{Spearman $\rho$}
        & \textbf{CoDrive Rank} \\
        \midrule
                 
        CD
        & UniDepth vs. VideoDepthAnything & 1.00 & \#1  \\

        RE
        & UniDepth vs. VideoDepthAnything & 0.94 & \#1  \\

        Hit
        & ByteTrack vs. BoT-SORT & 1.00 & \#1  \\

        LE
        & ByteTrack vs. BoT-SORT & 1.00 & \#1  \\
        
        \bottomrule
    \end{tabular}%
    }
\end{table*}

Since our consistency metrics rely on off-the-shelf perception components, we further examine whether the resulting model rankings are sensitive to the choice of evaluator. Specifically, we replace UniDepth~\citep{piccinelli2025unidepthv2} with VideoDepthAnything~\citep{chen2025videodepthanything} for the scene-consistency metrics and ByteTrack~\citep{zhang2022bytetrack} with BoT-SORT~\citep{aharon2022bot} for the instance-consistency metrics. As shown in Table~\ref{tab:evaluator-robustness}, the rankings remain highly stable across evaluator variants, with Spearman rank correlations of 1.00 for CD, 0.94 for RE, and 1.00 for both Hit and LE. Importantly, CoDrive remains ranked first under all evaluator configurations. These results suggest that the conclusions drawn from our benchmark are robust to reasonable changes in the underlying evaluation pipeline.

\section{Detailed Quantitative Results}

The results reported in the main text are aggregated across both the synthetic and real splits of the benchmark. Here, we present the individual results for each split separately. As shown in Tables~\ref{tab:quantitative-comparison-real} and~\ref{tab:quantitative-comparison-syn}, CoDrive achieves leading performance on both the real and synthetic sets.

\begin{table*}[h]
    \centering
    \caption{\textbf{Quantitative comparison of video generation methods on real-world data.}}
    \label{tab:quantitative-comparison-real}
    \resizebox{0.95\textwidth}{!}{%
    \begin{tabular}{lcccccccc}
        \toprule
        &
        & \multicolumn{2}{c}{\textbf{Controllability}}
        & \multicolumn{2}{c}{\textbf{Scene Consistency}}
        & \multicolumn{2}{c}{\textbf{Instance Consistency}} \\
        \cmidrule(lr){3-4}
        \cmidrule(lr){5-6}
        \cmidrule(lr){7-8}

        \textbf{Method}
        & \textbf{FVD} $\downarrow$
        & \textbf{ADE} $\downarrow$
        & \textbf{DTW} $\downarrow$
        & \textbf{CD} $\downarrow$
        & \textbf{RE} $\downarrow$
        & \textbf{Hit} $\uparrow$
        & \textbf{LE} $\downarrow$ \\
        \midrule
        
        HunyuanVideo-1.5
        & 882.3 & 11.20 & 599 & 2.04 & 0.53 & 0.527 & 8.87 \\
                
        Wan2.2-I2V
        & 850.5 & 27.13 & 1524 & 1.81 & 0.62 & 0.470 & 8.82 \\
        
        Cosmos3-Nano
        & 106.4 & 2.60 & 117 & 3.40 & 1.33 & 0.632 & 6.71 \\
        
        MagicDrive-V2
        & 2277.0 & 5.29 & 270 & 13.19 & 1.42 & 0.013 & 6.43 \\

        ShareVerse
        & 416.6 & 19.65 & 544 & 3.76 & 1.01 & 0.290 & 7.29 \\
        
        \midrule
        \textbf{CoDrive}
        & \textbf{104.3}
        & \textbf{1.53}
        & \textbf{61}
        & \textbf{1.57}
        & \textbf{0.45}
        & \textbf{0.646}
        & \textbf{5.99} \\

        \bottomrule
    \end{tabular}%
    }
\end{table*}

\begin{table*}[h]
    \centering
    \caption{\textbf{Quantitative comparison of video generation methods on synthetic data.}}
    \label{tab:quantitative-comparison-syn}
    \resizebox{0.95\textwidth}{!}{%
    \begin{tabular}{lcccccccc}
        \toprule
        &
        & \multicolumn{2}{c}{\textbf{Controllability}}
        & \multicolumn{2}{c}{\textbf{Scene Consistency}}
        & \multicolumn{2}{c}{\textbf{Instance Consistency}} \\
        \cmidrule(lr){3-4}
        \cmidrule(lr){5-6}
        \cmidrule(lr){7-8}

        \textbf{Method}
        & \textbf{FVD} $\downarrow$
        & \textbf{ADE} $\downarrow$
        & \textbf{DTW} $\downarrow$
        & \textbf{CD} $\downarrow$
        & \textbf{RE} $\downarrow$
        & \textbf{Hit} $\uparrow$
        & \textbf{LE} $\downarrow$ \\
        \midrule
        
        HunyuanVideo-1.5
        & 1676.8 & 12.71 & 548 & 2.11 & 0.69 & 0.118 & 12.97 \\
                
        Wan2.2-I2V
        & 283.0 & 18.77 & 855 & 1.82 & 0.66 & 0.134 & 13.49 \\
        
        Cosmos3-Nano
        & 205.8 & 8.78 & 366 & 3.85 & 1.22 & 0.115 & 12.96 \\
        
        MagicDrive-V2
        & 3468.9 & 16.1 & 822 & 25.86 & 1.13 & 0.001 & 19.57 \\

        ShareVerse
        & 240.0 & 10.86 & 254 & 4.66 & 1.20 & 0.027 & 11.51 \\
        
        \midrule
        \textbf{CoDrive}
        & \textbf{192.3}
        & \textbf{4.56}
        & \textbf{170}
        & \textbf{1.71}
        & \textbf{0.58}
        & \textbf{0.317}
        & \textbf{5.64} \\

        \bottomrule
    \end{tabular}%
    }
\end{table*}

\section{More Qualitative Results}\label{sec:appendix-qualitative-results}

\begin{figure*}[h]
    \centering
        \includegraphics[width=\linewidth]{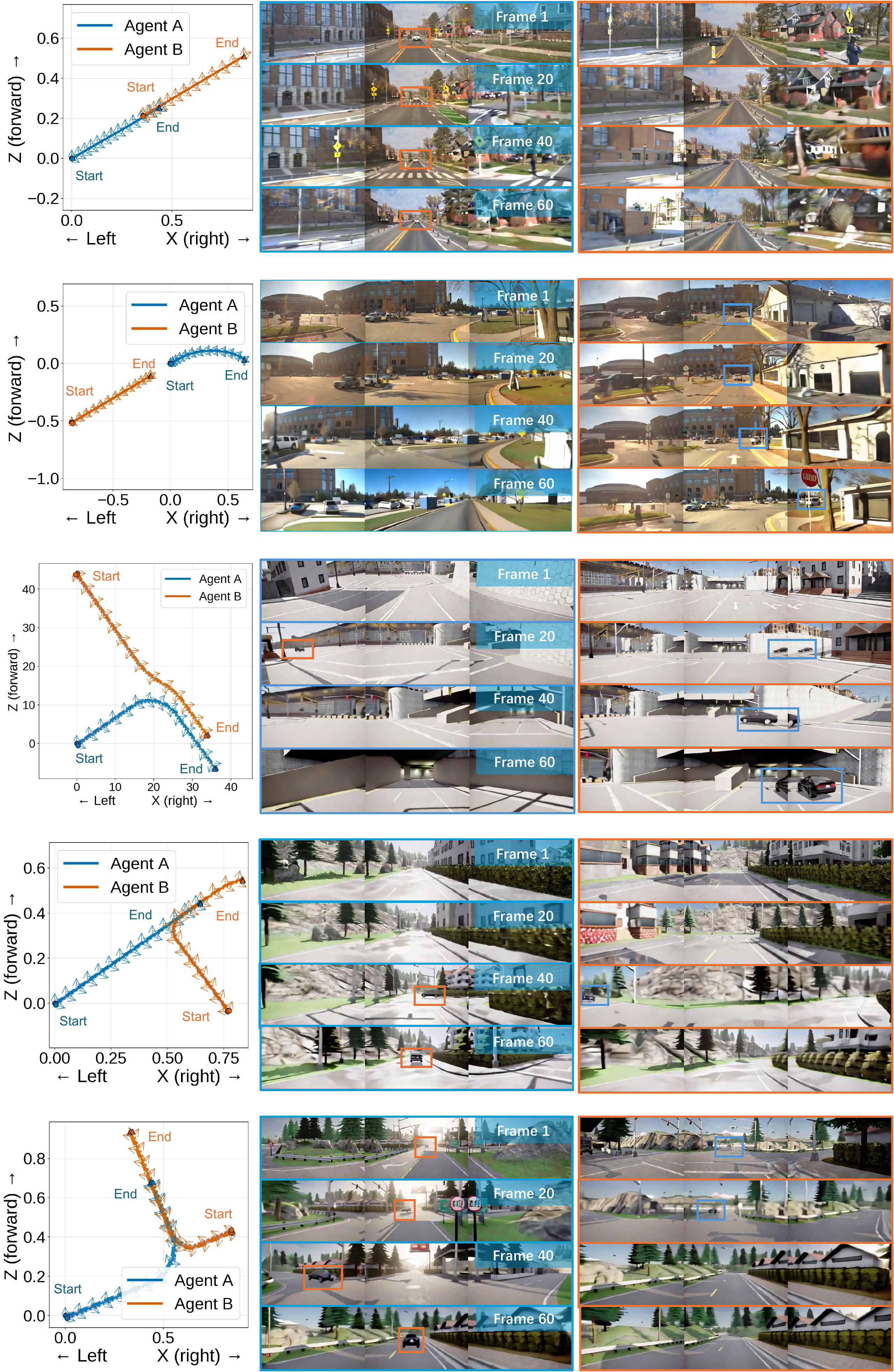}
    \caption{\textbf{Additional qualitative results}. Each row shows the relative trajectories of Agents A and B together with selected generated frames from both agents. }
\label{fig:more-qualitative}
\end{figure*}

Figure~\ref{fig:more-qualitative} presents additional qualitative examples with diverse relative vehicle trajectories. Each row shows the commanded top-down trajectories together with selected frames from the corresponding observations of Agents A and B, illustrating cross-agent scene and instance consistency over time.

\section{Limitations}\label{sec:appendix-limitation}

Our work still has several limitations: (1) Due to the scarcity and collection cost of synchronized real-world multi-vehicle data, cross-agent supervision in CoDrive currently comes from simulation. Although mixed-task fine-tuning improves performance on real-world scenarios, incorporating real-world cross-agent supervision remains important for reducing the simulation-to-real gap.
(2) PRoPE introduces an additional attention branch for geometric conditioning, increasing the computational cost of each transformer block in which it is applied.
(3) Although our formulation is written for $N$ vehicles, the current implementation and evaluation focus on two-vehicle interactions. Scaling joint generation and global attention to larger numbers of interacting vehicles remains an important direction for future work.